%% file: _000_main.tex
\pdfoutput=1

\documentclass[11pt]{article}

\usepackage[]{acl}

\usepackage{times}
\usepackage{latexsym}
\usepackage{enumitem}

\usepackage[T1]{fontenc}

\usepackage[utf8]{inputenc}

\usepackage{microtype}
\usepackage{multirow}
\usepackage{graphicx}
\newcommand{\longto}{\texttt{Longto}}
\newcommand{\longtonotes}{\texttt{LongtoNotes}}
\newcommand{\longtonotessmall}{$\texttt{LongtoNotes}_{s}$}
\newcommand{\longtonoteseq}{$\texttt{LongtoNotes}_{eq}$}

\usepackage{amssymb}
\usepackage{pifont}
\newcommand{\xmark}{\ding{55}}%
\usepackage{booktabs}

\usepackage{array}
\usepackage{graphicx}
\usepackage{multirow,makecell}
\title{\longtonotes: OntoNotes with Longer Coreference Chains}

\author{%
  Kumar Shridhar $^{\dag}$  \quad Nicholas Monath \thanks{\ \ Now at Google.} \ $^\ddag$   \quad Raghuveer Thirukovalluru$^\diamondsuit$ \\ 
  \bf{Alessandro Stolfo $^\dag$ \quad Manzil Zaheer  $^	\mathparagraph$ \quad Andrew McCallum $^\ddag$ \quad Mrinmaya Sachan $^\dag$} \\ \\
  $^{\dag}$ ETH Z{\"u}rich   \quad $^\ddag$ UMass Amherst  \quad $^\diamondsuit$ Duke University \quad $^\mathparagraph$ Google \\
  \texttt{shkumar@ethz.ch}
}

\begin{document}
\maketitle

\input{_010_abs}
\input{_020_intro}

\input{_030_corpus}
\input{_040_annotation_analysis}
\input{_050_empirical_analysis}

\input{_060_conclusions}
\input{_070_ethical_consideration}
\bibliography{anthology}

\input{08_appendix}

\end{document}

%% file: _010_abs.tex
\begin{abstract}
Ontonotes has served as the most important benchmark for coreference resolution. However, for ease of annotation, several long documents in Ontonotes were split into smaller parts.
In this work, we build a corpus of coreference-annotated documents 
of significantly longer length than what is currently available.
We do so by providing an
accurate, manually-curated, merging of 
annotations from documents that were split into multiple parts 
in the original Ontonotes annotation process \cite{ontonotes}.
The resulting corpus, which we call LongtoNotes contains documents in multiple genres of the English language with varying lengths, the longest of which are up to $8$x the length of documents in Ontonotes, and $2$x those in Litbank.
 We evaluate state-of-the-art neural coreference systems on this new corpus, analyze 
 the relationships between model architectures/hyperparameters and document length on
 performance and efficiency of the models, and demonstrate areas of improvement in long-document coreference modeling revealed by our new corpus. Our data and code is available at: \url{https://github.com/kumar-shridhar/LongtoNotes}.

\end{abstract}

%% file: _020_intro.tex
\section{Introduction}

Coreference resolution is an important problem in discourse
with applications in knowledge-base construction \cite{luan-etal-2018-multi}, question-answering
\cite{reddy-etal-2019-coqa} and reading assistants \cite{azab-etal-2013-nlp,head2021augmenting}.
In many such settings, the documents of interest, are significantly longer and/or on wider varieties of domains than the currently available corpora with coreference annotation \cite{ontonotes,litbank,mohan2019medmentions,cohen2017coreference}.
\begin{figure}[]
    \centering
    \includegraphics[width=0.48\textwidth]{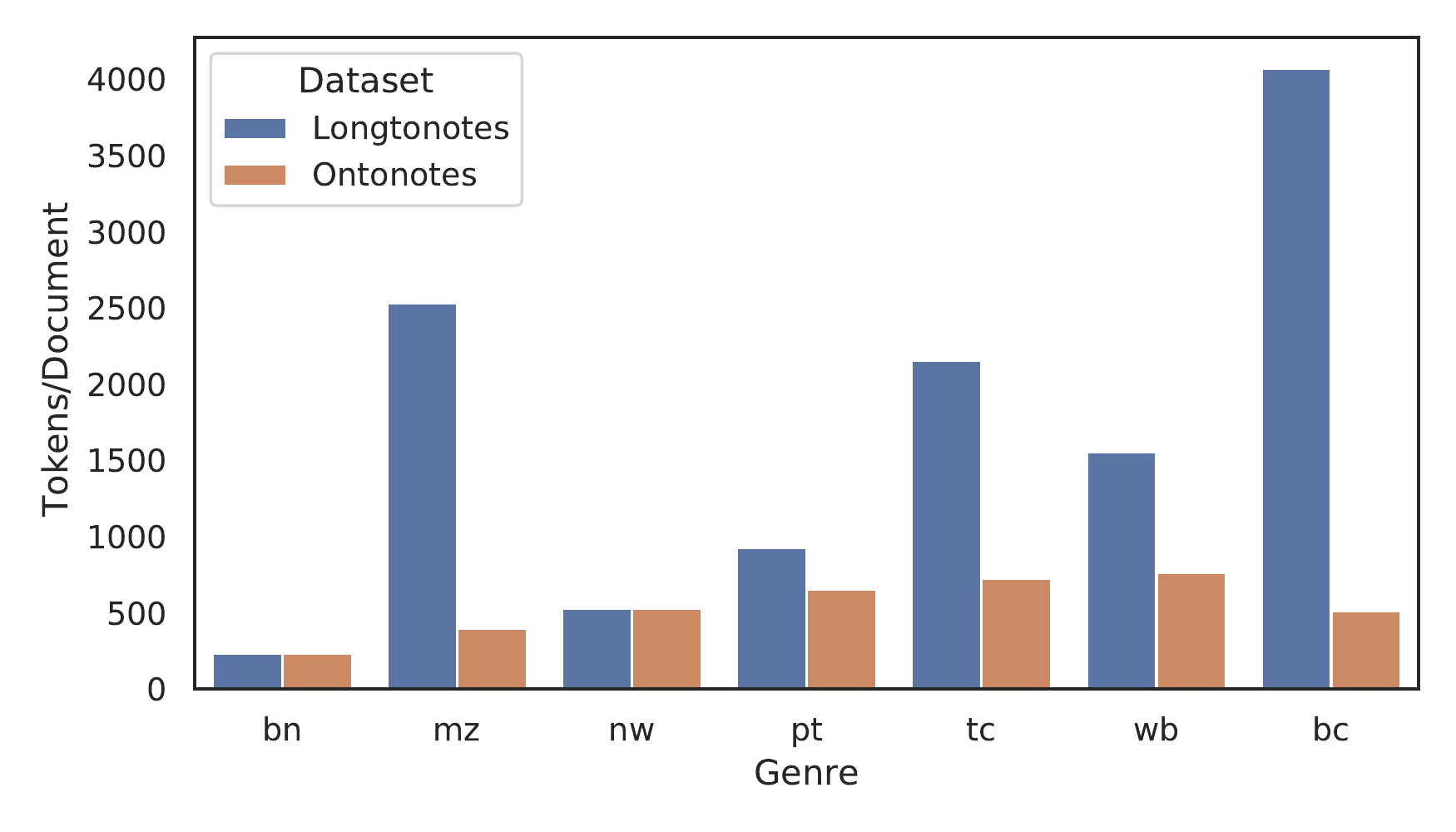}
    \caption{\textbf{Comparing Average Document Length}. Long documents in genres such as \emph{broadcast conversations (bc)} were split into smaller parts in Ontonotes. Our proposed dataset, \longtonotes, restore documents to their original form, revealing dramatic increases in length in certain genres. 
    }
    \label{tab:avg_doc_leng}
    \vspace{-4mm}
\end{figure}

The Ontonotes corpus \cite{ontonotes} is perhaps the most widely used benchmark for coreference \citep{lee2013deterministic,durrett2013easy,wiseman-etal-2016-learning,lee-etal-2017-end,joshi2020spanbert,toshniwal-etal-2020-learning, thirukovalluru-etal-2021-scaling, Kirstain2021CoreferenceRW}. The construction process for 
Ontonotes, however, resulted in documents with an artificially reduced length. For ease of annotation,
longer documents were split into smaller parts and each part was annotated separately and treated as an independent document \cite{ontonotes}. The result is a corpus in which certain genres, such as \emph{broadcast conversation (bc)}, have greatly reduced length compared to their original form (Figure~\ref{tab:avg_doc_leng}). As a result, the long, bursty spread of coreference chains in these documents is missing from the evaluation benchmark.

In this work, we present an extension to the Ontonotes corpus, called \longtonotes. \longtonotes\  combines coreference annotations in various parts of the same document, leading to a full document coreference annotation. This was done by our annotation team, which was 
carefully trained to follow the annotation guidelines laid out in the original Ontonotes corpus (\S\ref{sec:annotation}). This led to a dataset where the average document length 
is over 40\% longer than the standard OntoNotes benchmark and the average size of coreference chains increased by 25\%.
While other datasets such as Litbank \cite{litbank} and  CRAFT \cite{cohen2017coreference} focus on long documents in specialized domains, \longtonotes\ comprises of documents in multiple genres 
(Table~\ref{GenreCompar}).

To illustrate the usefulness of \longtonotes, we evaluate state-of-the-art coreference resolution models \cite{Kirstain2021CoreferenceRW,toshniwal-etal-2020-learning,joshi2020spanbert} on the corpus and analyze the performance in terms of document length (\S\ref{sec:length_empirical}). We illustrate how model architecture decisions and hyperparameters that support long-range dependencies have the greatest impact on coreference performance and importantly, these differences are only illustrated using \longtonotes\ and are not seen in Ontonotes (\S\ref{sec:hyperparameter_empirical}). \longtonotes\  also presents a challenge in scaling coreference models as prediction time and memory requirement increase substantially on the long documents~ (\S\ref{sec:efficiency_empirical}).

%% file: _030_corpus.tex
\vspace{-1mm}
\section{Our Contribution: \longtonotes}
\vspace{-1mm}

We present \longtonotes, a corpus that extends the English 
coreference annotation in the OntoNotes Release 5.0 corpus\footnote{The Arabic and Chinese parts of the Ontonotes dataset are not considered in our study. 
See Appendix~\ref{appendix:selection_discuss}} 
\cite{ontonotes} to provide annotations for longer documents. In the original English OntoNotes corpus, the genres such as 
\emph{broadcast conversations (bc)} and \emph{telephone conversation (tc)} contain long 
documents that were divided into smaller parts to facilitate
easier annotation. \longtonotes\ is constructed by collecting annotations 
to combine within-part coreference chains 
into coreference chains over the entire long document. The annotation procedure, in which annotators merge coreference chains, is described and analyzed in Section~\ref{sec:annotation}.

The divided parts of a long document in Ontonotes are all assigned to the same partition (train/dev/test). This allows \longtonotes\ to maintain the same train/dev/test partition, at the document level, as Ontonotes (Appendix, Table \ref{Train-test-devsplit}). The size of these partitions however does change as the divided parts are combined into a single annotated text in \longtonotes. We will release scripts to convert OntoNotes to \longtonotes\ in both CoNLL and CorefUD (Universal Dependencies)\footnote{\url{https://ufal.mff.cuni.cz/corefud}} formats under the Creative Commons 4.0 license.

We refer to \longtonotessmall\ as the subset of \longtonotes\ comprising only of long documents (i.e. documents merged by the annotators). 

\subsection{Length of Documents in \longtonotes}

The average number of tokens per document (rounded to the nearest integer) in \longtonotes\ is 674, ~44\% higher than in Ontonotes (466). Table~\ref{GenreCompar} breaks down the changes in document length by genre. We observe that the genre with the longest documents is \emph{broadcast conversation} with 4071 tokens per document, which is a dramatic increase from the length of the divided parts in Ontonotes which had 511 tokens per document in the same. The number of coreference chains and the number of mentions per chain grows as well. 
The long documents that were split into multiple parts during the original OntoNotes annotation are not evenly distributed among the genres of text present in the corpus.  In particular, text categories \emph{broadcast news (bn)} and \emph{newswire (nw)} consist exclusively of short non-split documents, which were not affected by the \longtonotes\  merging process.
A detailed distribution of what documents are merged in \longtonotes\ is provided in Table~\ref{DocComparison} in the Appendix.

\begin{figure}[t]
\vspace{-5mm}
\centering
\includegraphics[width=0.45\textwidth]{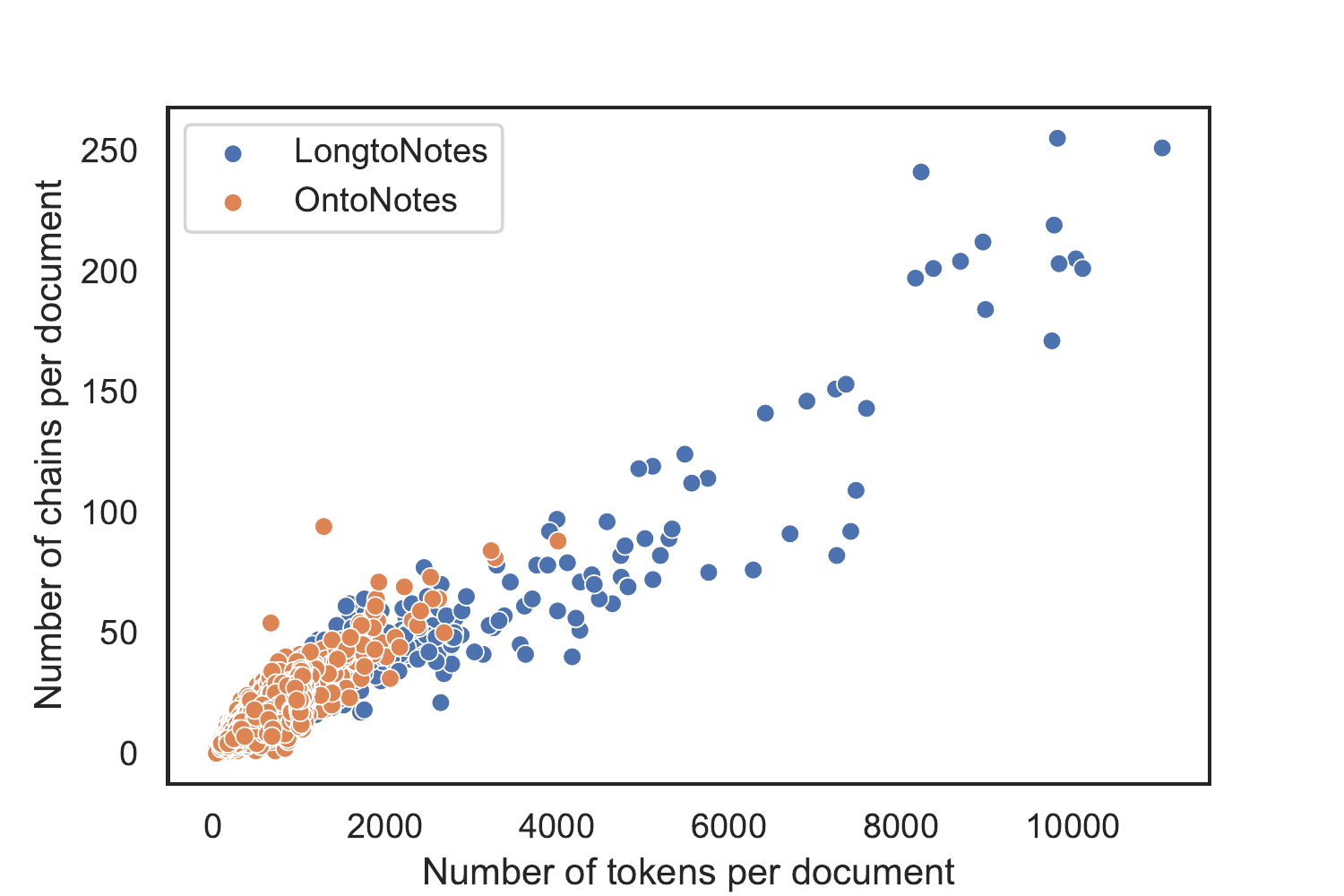}
\caption{\textbf{Document and Coref Chain Length.} The number of coreference chains increases with the increase in token length in \longtonotes.}
\label{tokensvscluste}
\end{figure}

\subsection{Number of Coreference Chains}
As a consequence of the increase in document length, \longtonotes\ presents a higher number of coreference chains per document (16), compared to OntoNotes (12). Figure~\ref{tokensvscluste} shows the length and number of coreference chains for each document in the two corpora. As expected, the number of chains in a document tends to get larger as the document size increases.

\begin{figure}[t]
\centering
\includegraphics[width=0.45\textwidth]{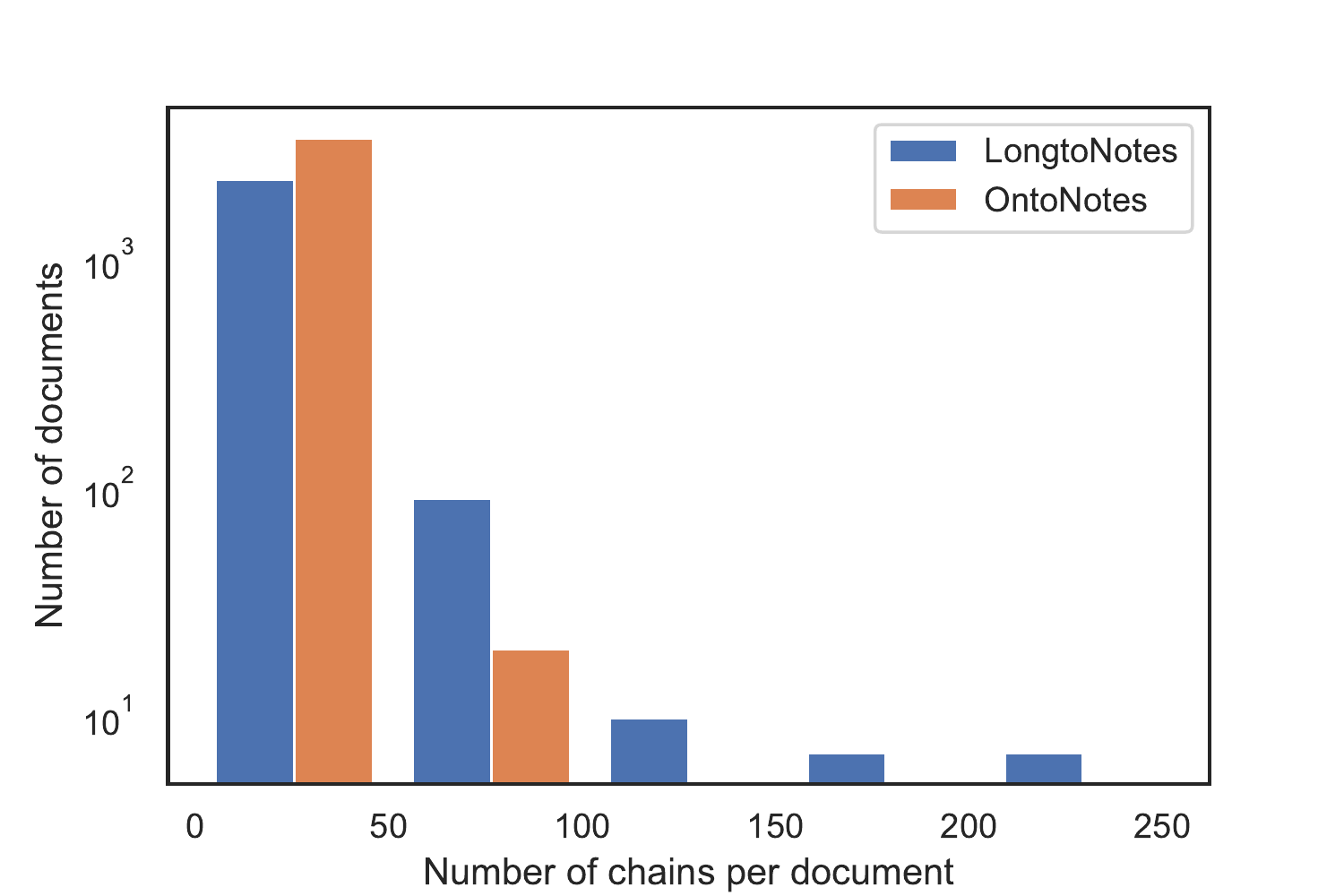}
\caption{\textbf{Number of Chains per Document.} A histogram log plot reveals the long-tailed distribution of the number of coreference chains present per document in \longtonotes. Ontonotes contain more documents with fewer chains.}
\label{histogram_clusters}
\end{figure}

For genres with longer average document lengths like \emph{broadcast conversation (bc)}, the increase in the number of chains is as high as $85\%$, while this increase is only $25\%$ for \emph{pivot (pt)} genre when the document length is comparatively shorter.  
It is worth noting that the majority of documents had a number of chains in the range of $20$ to $50$ and only about $20$ documents out of $3493$ in the OntoNotes dataset had $>$50 chains per document. For \longtonotes\, the number increases to $96$ documents. A comparison of the number of chains per document between OntoNotes and \longtonotes\ is shown in Figure \ref{histogram_clusters}. 

\vspace{-1mm}

\subsection{Number of Mentions per Chain}
The number of mentions per coreference chain in \longtonotes\ is over $30\%$ more than OntoNotes. This is primarily because of longer documents and an increase in the number of coreference chains per document. Mentions per chain increase with the increase in document length. For the \emph{broadcast conversation (bc)} genre, the increase in the mentions per chain is highest with $87\%$, while for the \emph{pivot (pt)} (Old Testament and New Testament text) genre it is only $30\%$ as it has shorter documents.

\subsection{Distances to the Antecedents}

\begin{figure}[h]
\centering
\includegraphics[width=0.5\textwidth]{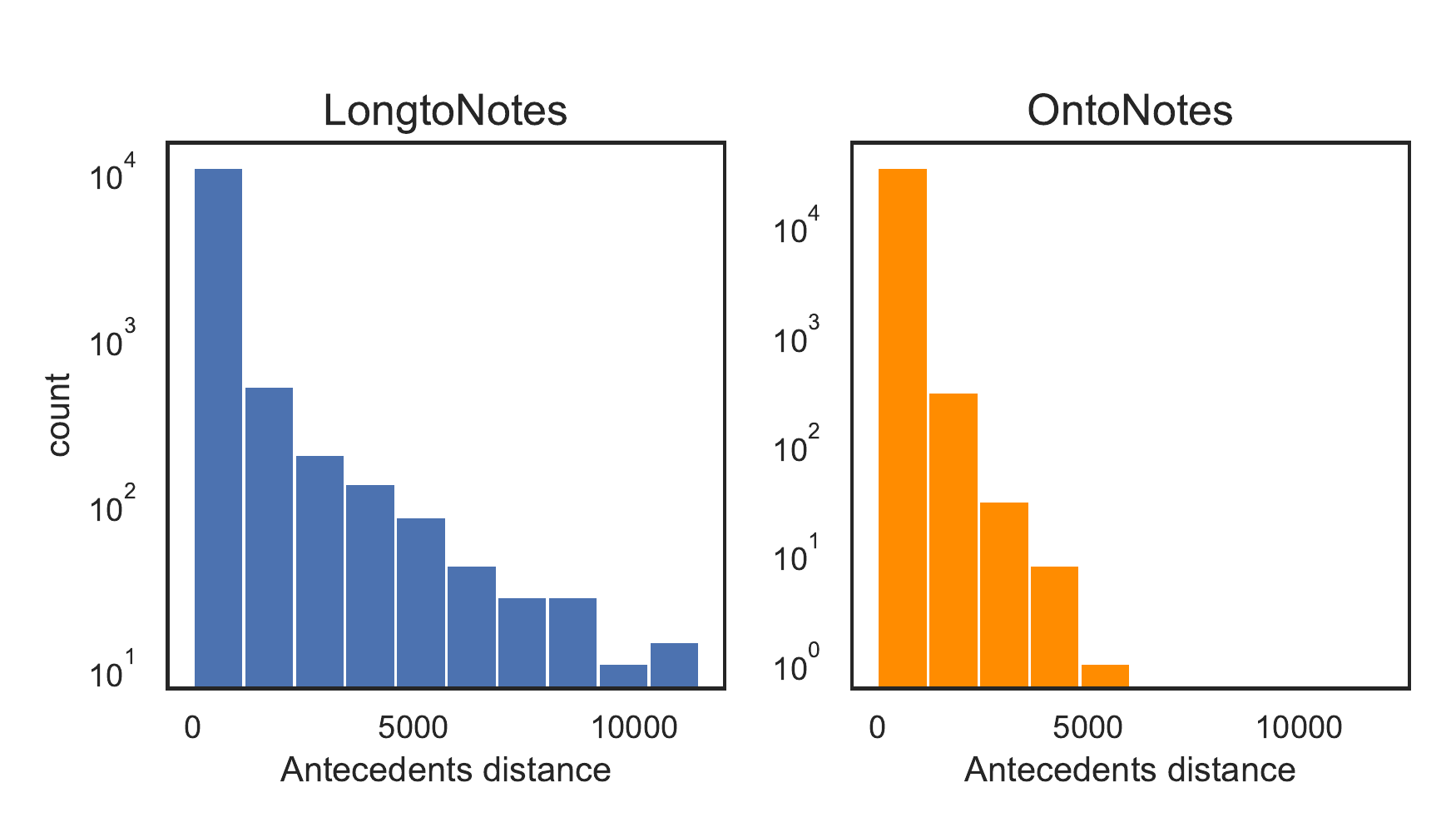}
\caption{\textbf{Distance to Antecedent}. Histogram (log-scale) shows that the largest distance of mention to their antecedents per chain increases in \longtonotes\ compared to OntoNotes.}
\label{hist_ment}
\end{figure}

For each coreference chain, we analyzed the distance between the mentions and their antecedents. The largest distance for a mention to its antecedent grew $3$x for \longtonotes\ dataset when compared to OntoNotes from 4,885 to 11,473 tokens. Figure \ref{hist_ment} shows a detailed breakdown of the mention to antecedent distance. There are no mentions that are more than $5$K tokens distant from its antecedent in OntoNotes. There are $178$ such mentions in \longtonotes.  

\vspace{-1mm}
\subsection{Comparison with other Datasets}

\begin{table*}[]
\centering\small
\begin{tabular}{l cc cc cc cc}
 \toprule
 \bf Categories  & \multicolumn{2}{c}{\bf \# Docs} & \multicolumn{2}{c}{\bf Tokens/Doc} & \multicolumn{2}{c}{\bf \# Chains} & \multicolumn{2}{c}{ \bf Ment./Chains} \\
  {} &  Ont. & Long. & Ont. & Long. & Ont. & Long. & Ont. & Long. \\
 \hline
 broadcast conversation (bc)    & 397 & 50 & 511 & 4071 & 14 & 85 & 65 & 519  \\
 broadcast news (bn)    & 947 &  947 & 237 & 237 & 8 & 8 & 29 & 29  \\
 magazine (mz)   & 494  & 78 & 398 & 2531 & 8 & 41 & 32 & 208 \\
 newswire (nw)   & 922  & 922 & 529 & 529 & 12 & 12 & 47 & 47 \\
 pivot (pt) & 369 &  261 & 657 & 930 & 20 & 27 & 131 & 186 \\
 telephone conversation (tc)  & 142 &  48 & 728 & 2157 & 17 & 44 & 108 & 319\\
 web data (wb)   & 222 & 109 & 763 & 1555 & 17 & 31 & 73 & 149 \\
 \hline
 Overall & 3493 & 2415 & 466 & 674 & 12 & 16 & 55 & 80\\
 \bottomrule
 \end{tabular}
 \caption{\textbf{Genre Comparison}. Comparison of document and coreference chain statistics per genre in OntoNotes 5.0 and our proposed dataset, \longtonotes.}
 \label{GenreCompar}

\end{table*}

The literature contains multiple works proposing datasets for coreference resolution: Wiki coref \cite{ghaddar-langlais-2016-wikicoref}, LitBank \cite{litbank}, PreCo \cite{preco}, Quiz Bowl Questions \citep{rodriguez2019quizbowl, guha-etal-2015-removing}, ACE corpus \cite{walker2006ace}, MUC \cite{chinchor1995message}, MedMentions \cite{mohan2019medmentions}, inter alia. We compare \longtonotes\ to these datasets in terms of the number of documents, the total number of tokens, and document length (Table \ref{DatasetComparison}).

Litbank is a popular long document coreference dataset, presenting a high tokens/document ratio. However, the dataset consists of only 100 documents, rendering model development challenges. Moreover, it focuses only on the literary domain. Other datasets containing long documents (e.g., WikiCoref) are also very small in size.
On the other hand, datasets consisting of a larger number of texts tend to contain shorter documents (e.g., PreCo). Thus, by building \longtonotes\ , we address the scarcity of a multi-genre corpus with a collection of long documents containing long-range coreference dependencies.

\begin{table}[h]
\small
\centering
\begin{tabular}{l c c c}
 \toprule
 \bf Dataset  & \# Docs & Total Size & Tokens/Doc\\
 \hline
 WikiCoref    & 30 &  60K & 2000\\
 ACE-2007   & 599  & 300K & 500\\
 MUC-6   & 60  & 30K & 500\\
 MUC-7   & 50  & 25K & 500\\
 QuizBowl & 400 &  50K & 125\\
 PreCo  & 37.6K & 12.4M & 330\\
 LitBank   & 100  & 200K & 2105\\
 MedMentions & 4392 & 1.1M & 267 \\
 OntoNotes   & 3493 & 1.6M & 466\\
 \hline
 \longtonotes   & 2415 & 1.6M & 674\\
 \longtonotessmall   & 283 & 740K & 2615 \\
 \bottomrule
 \end{tabular}
 \caption{\textbf{Coreference Datasets}. A comparison of various coref datasets with our proposed dataset \longtonotes.}
 \label{DatasetComparison}

\end{table}

%% file: _040_annotation_analysis.tex
\section{Annotation Procedure \& Quality}

\label{sec:annotation}

In this section, we describe and assess the annotation procedure
used to build \longtonotes.

\subsection{Annotation Task}

To build \longtonotes\ , it suffices to successively merge chains in the current part $i+1$ of the document with one of the chains in the previous parts $1,\dots,i$. 

We reformulate this annotation process as a question-answering task where we ask annotators a series of questions (rather the same coreference determining question for different mentions) using our own annotation tool designed for this task (Appendix, Figure~\ref{tool}). We display parts $1,\dots,i$ with color-coded mention spans. We then show a highlighted concept (a co-reference chain in part $i+1$) and ask the question: \emph{The highlighted concept below refers to which concept in the above paragraphs?}. The annotators select one of the color-coded chains from parts $1,\dots,i$ from a list of answers or the annotators can specify that the highlighted concept in part $i+1$ does not refer to any concept in parts $1,\dots,i$, (i.e., a new chain emerging in part $i+1$). 

The annotation tool proceeds with a question for each coreference chain ordered (sorted by the first token offset of the first mention in the chain).

The annotation of all parts of a document comprises an annotation task. That is, a single annotator is tasked with answering the multiple-choice question for each coreference chain in each part of a document. 

At the end of each part, annotators are shown a summary page that allows them to review, modify, and confirm the decisions made in the considered part.
A screenshot of the summary page is provided in Fig. \ref{tool-summary} in the Appendix.

\paragraph{From Annotations to Coreference Labels}
The annotations collected in this way are then converted into coreference labels for the merged parts of a document. The answers to the questions tell us the antecedent link between two coreference chains. These links are used to relabel all mentions in the two chains with the same coreference label, resulting in the \longtonotes\ dataset.

\paragraph{Annotation of singletons}
Note that the existing OntoNotes coreference annotation does not include singletons.
 
However, considering all parts of a document together might allow mentions that were considered to be singletons in a specific part to be assigned to a coreference chain.
To understand the frequency of singletons in a single part of a document that has coreferent mentions in other parts, we manually analyzed 500 mentions spread across 10 parts over three randomly selected long documents. We found only 17 instances ($\sim 0.03\%$) where singletons can be merged with coreference chains in different parts of the same document. Given that such singletons would constitute only such a small percentage of mentions, we decided it was appropriate to leave them out of the annotation process to reduce the complexity of the annotation task. To merge this small amount of singleton mentions, our annotators would have had to label over $50\%$ more mentions per document. We further discuss this in Appendix \ref{appendix:singletons_discuss}.

\subsection{Annotators and Training}

We hired and trained a team of three annotators for the aforementioned task. 
The annotators were university-level English majors from India and were closely supervised by an expert with experience in similar annotation projects. The annotation team was paid a fair wage of approximately 15 USD per hour for the work.
We had several hour-long training sessions outlining the annotation task, setup of the problem, and Ontonotes annotation guidelines. We reviewed example cases of difficult annotation decisions and collaboratively worked through example annotations. We then ran a pilot annotation study with a small number of documents (approx 5\% of the total documents). 
For these documents, we also provided annotations to ensure the training of the annotators and eventual annotation quality. We calculated the inter-annotators’ agreement between the annotators and us. After a few rounds of training, we were able to achieve an inter-annotator agreement score (strict match, defined in the next subsection) of over 95\% and we decided to go ahead with the annotation task. This confirmed the annotators' understanding of the task.

\vspace{-0.04cm}
After the satisfactory pilot annotation study, the tasks were assigned to the annotators in five batches of 60 documents each. For 10\% of the tasks, we had all three annotators provide annotations. For the remaining 90\%, a single annotator 
was used. For the documents with multiple annotators, we used majority voting to settle disagreements. If all annotators disagreed on a specific case, we selected Annotator 1's decision over the others (analysis in the Appendix).

\noindent \textbf{Did we need annotation? Can the chains be merged automatically?}
To show the importance of our human-based annotation process, we investigate whether the annotators' decisions could have been replicated using off-the-shelf automatic tools. 
We performed two experiments: (i) a simple greedy rule-based string matching system (described in the Appendix \ref{appendix:greedy_discuss}) and (ii) Stanford rule-based coreference system to merge chains across various parts. We use the merged chains to calculate the CoNLL $F_1$ score with the annotations produced by our annotators. We found that our string-matching system achieved a CoNLL $F_1$ score of only $61\%$, while the Stanford coreference system reached a score of only $69\%$. The low scores compared to the annotators' agreement (which is over $90\%$) underline the complexity of the task and the need for such a human-annotated dataset.

\subsection{Measuring Quality of Annotation}

We would like to ensure that \longtonotes\ 
maintains the
high-quality standards of OntoNotes.

Thus, we compute various 
metrics of agreement between a pair of annotators. 
We consider (1) the question-answering agreement (i.e., how similar are the annotations made using the annotation tool), and (2) the coreference label agreement (i.e.,  at the level of the resulting coreference annotation).

Assume that each annotator receives a set of chains $C_1, C_2,..., C_N$. For each chain $C_i$, the annotator links it to a \texttt{New} chain or a chain from their (annotator specific) set of available chains. Let us call $D_i$ this linking decision, which consists of a pair $(C_i, A_i)$, where $A_i$ is the selected antecedent chain. We consider the following question-answering metrics: 

\textbf{(i) Strict Decision Matching}: When two annotators agreed on merging two chains and there is an exact match between the merged chains. Calculated as  $\frac{1}{N} \sum_i D_i^{(1)} = D_i^{(2)}$.
\newline

\textbf{(ii) Jaccard Decision Match}: Jaccard decision calculated as $\frac{1}{N} \sum_i \frac{(D_i^{(1)}.A_i^{(1)}) \cap (D_i^{(2)}.A_i^{(2)})}{(D_i^{(1)}.A_i^{(1)}) \cup (D_i^{(2)}.A_i^{(2)})}$
\newline

\textbf{(iii) New Chain Agreement}: Number of times two annotators agreed on a new chain choice divided by the number of times at least one annotator labels \textit{New} chain.
\newline

\textbf{(iv) Not New Chain Agreement}: Pairwise agreement between annotators when the chain choice is not a \textit{New} chain.
\newline

\textbf{(v) Krippendorff's alpha}: Krippendorff's alpha \cite{krippendorff2011computing} is the reliability coefficient measuring inter-annotator agreement.
We compute Krippendorff's alpha using a strict decision match as the coding for agreement.

\begin{table}[h]
\centering
\begin{tabular}{@{}lc}
    \toprule
    \textbf{Metric} &  \textbf{Score} \\
    \midrule
    \textbf{Strict Match}  & 0.90  \\\hline
    \textbf{Jaccard Match}  & 0.95 \\\hline
    \textbf{New Chain}  & 0.88  \\\hline
    \textbf{Not New Chain}  & 0.87  \\\hline
    \textbf{Krippendorff's alpha} & 0.90  \\
    \bottomrule
\end{tabular}
\caption{\textbf{Annotation Quality Assessment}. We report the average of each metric over all pairs of annotators. }
 \label{agreementMatrix}
\end{table}

Table \ref{agreementMatrix} presents the results for these metrics.
We observed that on average annotators agreed with each other on over $90\%$ of their decisions except when the \textit{No New} chains were considered. Removing \textit{New} chains reduces the total decisions to be made significantly, and hence a lower score on \textit{No New} chains agreement. 

We found that Annotator 1 agreed most with the experts and hence Annotator 1's decisions were preferred over the others in case of disagreement between all three annotators.

\paragraph{Where are disagreements found in annotation?}
We would like to understand what kinds of mentions lead to
the disagreement between annotators. To investigate this, we measure the part of speech of all the disagreed chain assignments between the annotators. We found that the $8\%$ of the mentions within the disagreed chain assignments were pronouns, $8\%$ were verbs, and $9\%$ were common nouns. The number of proper nouns disagreements was lower with just $5\%$. When considering different genres, it was observed that genres with longer documents like \emph{broadcast conversation (bc)} had more mentions that were pronouns when compared with genres with shorter documents \emph{pivot (pt)}. As expected, the number of disagreements in general increased with the size of the documents. However, we found that the number of disagreements was manageably small even for long document genres such as \emph{broadcast conversation (bc)}
A more comprehensive overlook is presented in the Appendix.

\subsection{Time Taken per Annotation}

We also recorded the time taken for each annotation.  Time taken per annotation increases with the increase in the document length (Appendix Fig. \ref{appendix:TimeAnnotatio}). This is expected as more chains create more options to be chosen from and longer document length demands more reading and attention. 

In total, our annotation process took $400$ hours. 

%% file: _050_empirical_analysis.tex
\section{Empirical Analysis with \longtonotes}

We hope to show that \longtonotes\ can facilitate 
the empirical analysis of coreference models in ways that were not possible with the original OntoNotes. 
We are interested in the following empirical 
questions using the datasets-- Ontonotes \cite{ontonotes}, and our proposed \longtonotes\ and \longtonotessmall:
\begin{itemize}
    \item How does the length of documents play a role in the empirical performance of models?
    \item Does the empirical accuracy of models depend on different hyperparameters in \longtonotes\ and Ontonotes?
    \item Does \longtonotes\ reveal properties about the efficiency/scalability of models not present in Ontonotes?
\end{itemize}

 \subsection{Models}

Much of the recent work on coreference can be organized into three categories: span based representations \citep{lee-etal-2017-end, joshi2020spanbert}, token-wise representations \citep{thirukovalluru-etal-2021-scaling, Kirstain2021CoreferenceRW} and memory networks / incremental models \citep{toshniwal-etal-2020-learning, toshniwal2020petra}. We consider one approach from all three categories.

\paragraph{Span based representation} We used the \citet{joshi2020spanbert} implementation of the higher-order coref resolution model \cite{lee2018higher} with SpanBERT. Here, the documents were divided into a non-overlapping segment length of 384 tokens. We used SpanBERT Base as our model due to memory constraints. 

The number of training sentences was set to $3$. We set the maximum top antecedents, $K = 50$. We used Adam \cite{kingma2014adam} as our optimiser with a learning rate of $2e^{-4}$. 

\paragraph{Token-wise representation} We used the LongFormer Large \cite{beltagy2020longformer} version of \citet{Kirstain2021CoreferenceRW} work, as this approach is less memory demanding and it is possible to fit this model in our memory. The max sequence length was set to $384$ or $4096$. Adam was used as an optimiser with a learning rate of $1e^{-5}$. A dropout \cite{JMLR:v15:srivastava14a} probability of $0.3$ was used. 

\paragraph{Memory networks} We used SpanBERT Large with a sequence length of $512$ tokens. Following \citet{toshniwal-etal-2020-learning}, an endpoint-based mention detector was trained first and then was used for coreference resolution. The number of training sentences was set to $5$, $10$, and $20$. The number of memory cells was selected from $20$ or $40$. All experiments were performed with AutoMemory models with learned memory type.

\subsection{Length of Documents \& Performance}
\label{sec:length_empirical}

\paragraph{Impact of Training Corpus}
We first investigate whether or not training on the longer
documents in \longtonotes\ are needed to achieve
state-of-the-art results on the dataset. 
We compare the performance of models trained on Ontonotes 
to those trained on \longtonotes. We find that by training on \longtonotes, we can achieve higher CoNLL F1 measures
on \longtonotes\ than training with Ontonotes for each model architecture (Table~\ref{ComparF1}).

This suggests that the longer dependencies formed by merging annotations in various parts of documents in OntoNotes are difficult to model when training on short documents.

We find that to achieve accuracy with hyperparameters such as learning rate/warmup size, we need to maintain a number of steps per epoch consistent with Ontonotes when training with \longtonotes. A detailed analysis is presented in the Appendix Section~\ref{appendix:empirical}. 

\begin{table}[t]
\small
\centering
\begin{tabular}{l l c}
\toprule
\bf \# Tokens & \bf Training & \bf CoNLL F1 \\
\midrule
\multirowcell{2}[0pt][l]{$\leq2$K} & Ontonotes & \textbf{78.85}  \\
&\longtonotes & 78.25  \\
\midrule
\multirowcell{2}[0pt][l]{$>2$K} & Ontonotes & { 65.11}  \\
&\longtonotes & \textbf{66.20}  \\

 \bottomrule
 \end{tabular}
 \caption{\textbf{Performance and Document Length for Span-based Models}.$F_1$ score across different document length for SpanBERT Base trained model on OntoNotes and \longtonotes\ dataset.}
 \label{F1vsDoc}
\end{table}

\begin{table*}
\resizebox{\linewidth}{!}{
\begin{tabular}{@{}l l  ccc  ccc  ccc}
 \toprule
 \bf &   & \multicolumn{3}{c}{\bf OntoNotes} & \multicolumn{3}{c}{\bf \longtonotessmall} & \multicolumn{3}{c}{\bf \longtonotes} \\

  {} & \bf Training &  P & R & $F_1$ & P & R & $F_1$ & P & R & $F_1$ \\
 \midrule
Stanford Coref \cite{lee2013deterministic} & - &  58.6 & 58.8 & 58.6 & 48.5 & 58.2 & 52.7 & 53.6 & 57.3  & 55.2\\
\midrule
\multirowcell{2}[0pt][l]{Span-based \\ \cite{joshi2020spanbert}} & OntoNotes & 76.5 & 77.6 & \textbf{77.4} & 72.7 & 69.1 & 70.8 & 74.4 & 73.0  & 73.7\\

& \longtonotes   & 75.9 & 77.7 & 76.8 & 72.4 & 70.7 & \textbf{71.5} & 73.9 & 74.1 & \textbf{74.0}\\
\midrule
 \multirowcell{2}[0pt][l]{Token-Level \\ \cite{Kirstain2021CoreferenceRW}} & Ontonotes & 81.2  & 79.5 & \textbf{80.4} & 79.6 & 80.0 & 79.8 & 79.7 & 77.2 & 78.5 \\
 & \longtonotes & 80.0  & 78.2 & 79.1 &  80.3 & 80.3 & \textbf{80.3}  & 80.2 & 78.0 & \textbf{79.1} \\
 \midrule
 \multirowcell{2}[0pt][l]{Memory-Model \\ \cite{toshniwal-etal-2020-learning}} & OntoNotes   & 73.5  & 79.3 & 76.4 & 63.4 & 73.8 & 68.2 & 67.9 & 76.6 & 72.0  \\

 & \longtonotes   & 73.8 & 79.4 & \textbf{76.6} & 66.3 & 74.6 & \textbf{70.2} & 69.3 & 77.0 & \textbf{72.9}\\

 \bottomrule
 \end{tabular}
 }
 \caption{\textbf{Performance Variation by Training Set}. Comparison of \textbf{$F_1$} scores on various datasets using different models. {\bf All experiments have been performed atleast 2 times and a variance of only $\pm\ 0.1$ was observed.}} 
 \label{ComparF1}
\end{table*}

\paragraph{Length Analysis - Number of Tokens} 
We break down the performance of the Span-based model by the number of tokens in each document. We compare the performance of the model depending on the training set. 
Figure \ref{tokensvscluste} shows that the majority of the documents in the OntoNotes dataset falls within a token length of $2000$ per document. We create two splits of \longtonotessmall, one having a token length greater than $2000$ tokens, the other having a number of tokens smaller than $2000$. Table \ref{F1vsDoc} shows that for smaller document length (less than $2000$ tokens), the SpanBERT model trained on OntoNotes performed better but the trend reverses for longer documents (more than $2000$ tokens), on which the model trained on \longtonotes\ outperformed the model trained on OntoNotes by $+1\%$.

\paragraph{Length Analysis - Number of Clusters} 
Table \ref{F1vsCluster} displays the change in $F_1$ score with the increase in the number of clusters per document. The SpanBERT Base model trained on \longtonotes\ outperforms the same model trained on OntoNotes ($+0.6\%$) when the number of clusters is more than $40$. Note that, $40$ is selected based on the cluster distribution shown in Table \ref{GenreCompar} with the majority documents in \longtonotes\ lying in this range.

\subsection{Hyperparameters \& Document Length}
\label{sec:hyperparameter_empirical}

Each model has a set of hyperparameters that would seemingly
lead to variation in performance with respect to document 
length. We consider the performance of the models on \longtonotes\
as a function of these hyperparameters.

\begin{table}[t]
\small
\centering
\begin{tabular}{l l c c c}
\toprule
\bf \# Chains & \bf Training & \bf SpanBERT & \bf Token & \bf Memory\\
\midrule
\multirowcell{2}[0pt][l]{$\leq40$} & Onto &  \textbf{73.60} &\textbf{79.80}  &\textbf{72.80} \\
& \longto & 72.86  & 78.80 & 71.94  \\
\midrule
\multirowcell{2}[0pt][l]{$>40$} & Onto &  68.44   & 75.60 & 67.72 \\
&\longto &   \textbf{69.09}  & \textbf{76.42} & \textbf{68.60}  \\
 \bottomrule
 \end{tabular}
 \caption{\textbf{Performance and Number of Chains for different models}. CoNLL $F_1$ score across different document lengths for SpanBERT Base, Token-Level, and Memory-Model trained on OntoNotes and \longtonotes\ dataset.}
 \label{F1vsCluster}
\end{table}

\paragraph{Span-based model hyperparameters} We consider two 
hyperparameters: the number of antecedents to use, $K$
and the max number of sentences used in each training 
example. We found that upon varying $K$: $10, 25 $ and $50$, there was only a small difference observed in the results for both the models trained on OntoNotes and \longtonotes\ (increasing K led to only minor increases). The result is summarized in Table \ref{VaryK}. We could not go beyond $K=50$ due to our GPU memory limitations. However, going beyond $50$ might further help for longer documents. 

\begin{figure}[h]
\centering
\includegraphics[width=0.5\textwidth]{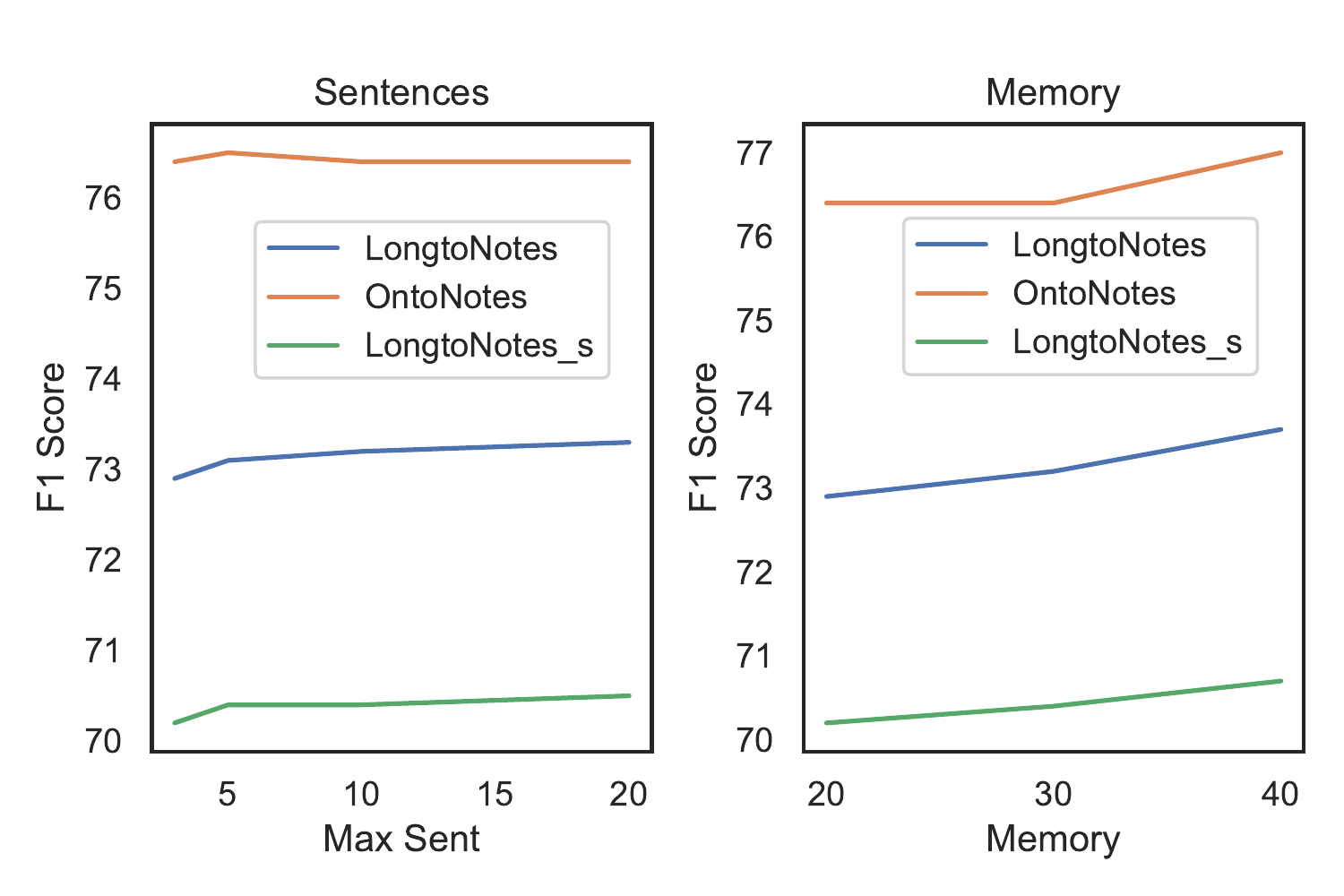}
\caption{\textbf{Max Sentence Length.} Increasing max sentences from $3$ to $20$ has a small effect on the performance of the SpanBERT large model. On the other hand, the increase is linear with the increase in the memory size alongside the increase in max training sentences.}
\label{sentvsmem}
\end{figure}

Furthermore, we found that the \textit{number
of sentences} parameter used to create training batches does not play a significant role in performance either (Figure \ref{sentvsmem}).

\begin{table}[h!]
\small
\centering
\begin{tabular}{l  c c c}
 \toprule
  \bf K & \bf OntoNotes & \bf \longtonotes & \bf \longtonotessmall \\
 \midrule
  10 & 77.05 &  73.44 & 70.37 \\
  25 & 76.93 &  73.99 & \textbf{71.61} \\
  50 & \textbf{77.60} &  \textbf{74.01} & 71.58 \\
 \bottomrule
 \end{tabular}
 \caption{\textbf{Number of Antecedents vs. Performance} SpanBERT Base model trained on \longtonotes\ dataset with varying $K$ value.} 

 \label{VaryK}
\end{table}

\begin{figure}[h]
\centering
\includegraphics[width=0.5\textwidth]{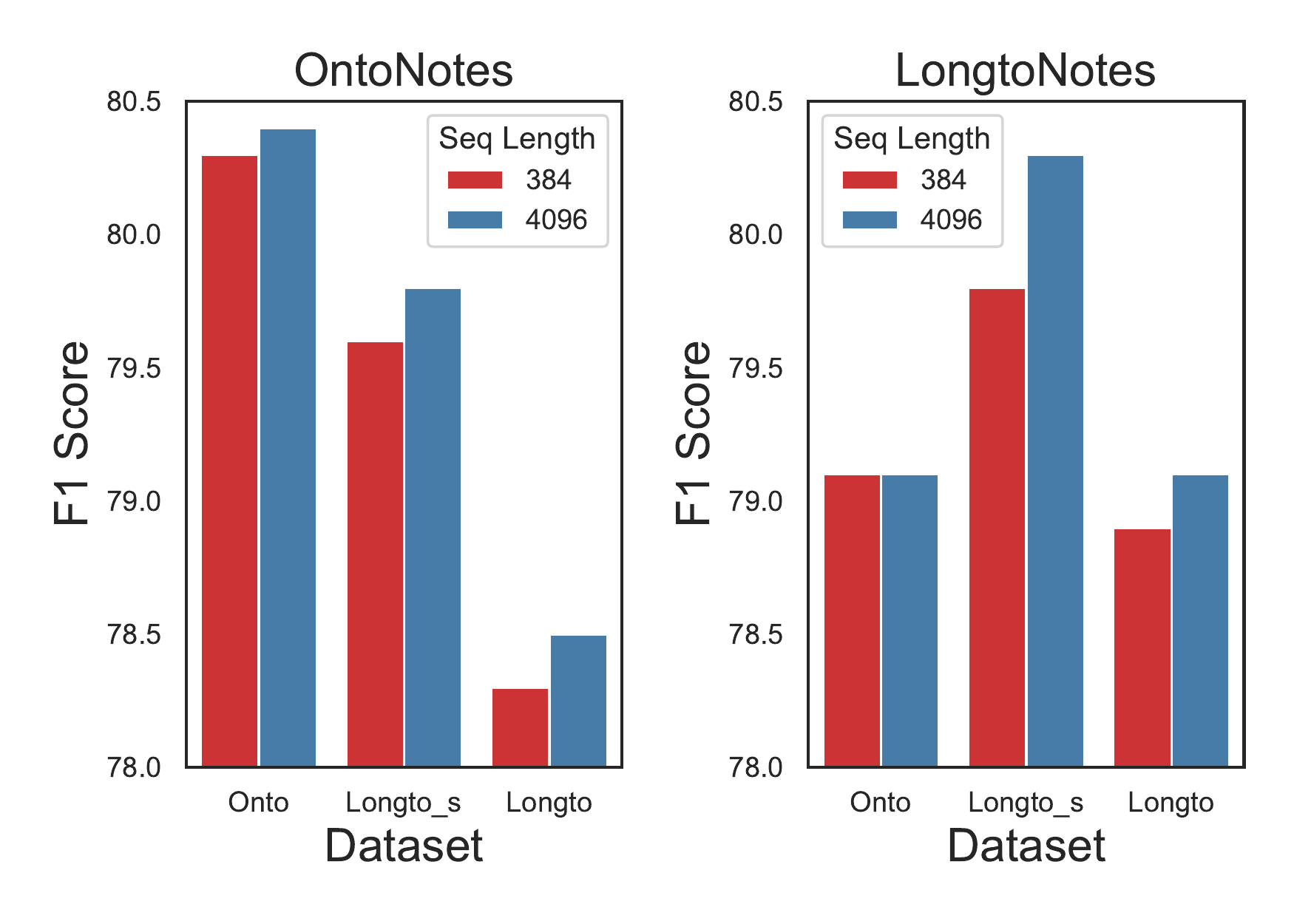}
\caption{\textbf{Sequence Length vs. Performance.} LongFormer is significantly better on \longtonotes\ with 4096 sequence length compared to 384. Two sequence lengths perform similarly on Ontonotes.} 
\vspace{-1mm}
\label{384vs4096}
\end{figure}

\vspace{-0.5cm}
\paragraph{Token-wise model hyperparameters}
We experimented with reducing the sequence length when testing from $4096$ to $384$ and we observe a drop in performance. Figure \ref{384vs4096} shows the effect on performance due to the change in the sequence length. We observed that longer sequence length ($4096$) helps more for \longtonotessmall\ as there are longer sequences than for OntoNotes, which is evident in Figure \ref{384vs4096}. Furthermore, we analyzed the effect of sequence length on two genres: \emph{magazine (mz)} having $6$x longer sequences in \longtonotes\ than OntoNotes vs \emph{pivot (pt)} having just $1.4$x longer documents. As observed in Figure \ref{genreF1}, when the document is long as in \emph{magazine (mz)}, there is a significant increase in performance with a longer sequence but the effect is negligible for \emph{pivot (pt)} where the size of the document is almost the same. A detailed comparison is provided in the Appendix Table \ref{ComparF1genre384}.

\paragraph{Memory model hyperparameters} We consider
two hyperparameters - the memory size which denotes the maximum active antecedents that can be considered and the max number of sentences
used in training. We show that doubling the size of the memory leads to an increase of $0.8$ points of CoNLL $F_1$ for \longtonotes\ dataset. (Appendix Table~\ref{memorysizeVF1}).
Figure \ref{sentvsmem} demonstrates that there is no significant improvement in the performance of the model with the increase in the number of training sentences.

\subsection{Model Efficiency}
\label{sec:efficiency_empirical}

We compare the prediction time for the span-based
model on the longest length and average length documents in \longtonotes\ and Ontonotes in Table~\ref{tab:eff}. We observe that there is a significant jump in running time and memory required to scale the model to long documents on \longtonotes; this jump is much smaller on Ontonotes.
This suggests that our proposed dataset is better suited for assessing the scaling properties of coreference methods.

\begin{table}[h]
    \small 
    \centering
    \begin{tabular}{lccc}
        \toprule
        \bf Dataset & \bf Type & \bf Pred. Time & \bf Pred. Mem \\
        \midrule
        Ontonotes & Average & 0.11 sec & 1.50 GB\\
        \longtonotes & Average & 0.47 sec & 6.50 GB\\
        Ontonotes & Longest & 0.37 sec  & 5.84 GB\\
         \longtonotes & Longest & 2.35 sec & 42.68 GB  \\
         \bottomrule
    \end{tabular}
    \caption{\textbf{Model Efficiency of Span-based Models}. We find that \longtonotes\ documents have extended length leading to greater variation of prediction time and prediction memory.}
    \label{tab:eff}
\end{table}

%% file: _060_conclusions.tex
\section{Conclusion}

In this paper, we introduced \longtonotes, a dataset that merges the coreference annotation of documents that in the original OntoNotes dataset were split into multiple independently-annotated parts. 
\longtonotes\ has longer documents and coreference chains than the original OntoNotes dataset. 
Using \longtonotes, we demonstrate that scaling current approaches to long documents has significant challenges both in terms of achieving  better performance as well as scalability. We demonstrate the merits of using \longtonotes\ as an evaluation benchmark for coreference resolution and encourage future work to do so.

%% file: _070_ethical_consideration.tex
\section*{Ethical Considerations}
Our dataset is comprised solely of English texts, and our analysis, therefore, applies uniquely to the English language.
The annotation was performed with a data annotation service which ensured that the annotators were paid fair compensation of 15 USD per hour. The annotation process did not solicit any
sensitive information from the annotators.
Finally, while our models are not tuned for any specific real-world application, the methods could be used in
sensitive contexts such as legal or health-care settings, and any work building on our
methods must undertake extensive quality-assurance and
robustness testing before using them.

\section*{Acknowledgements}

We acknowledge support from an ETH Zürich Research grant (ETH-19 21-1) and a grant from the Swiss National Science Foundation (project \# 201009) for this work.
This material is based upon work supported in part by the University of Massachusetts Amherst Center for Data Science and the Center for Intelligent Information Retrieval, and in part by the Chan Zuckerberg Initiative under the project ``Scientific Knowledge Base Construction''. We also thank the annotators (part of Xsaras.com, a data annotation company) for their sincere efforts in making this project possible.

%% file: 08_appendix.tex
\appendix
\section*{Appendix}
\section{Dataset and Annotation Details}
\label{sec:appendix}
\subsection{Annotation tool}
\begin{figure*}[t]
\caption{The tool designed by us for the annotation task. The upper box represents all the previous paragraphs while the box on the bottom left is the current paragraph. The mentions of the current chain to be merged are shown in yellow. On the right side, the answers are presented which are chains from previous paragraphs and the annotator can select one of them or choose the \texttt{None of the below} option which creates a new chain. }
\includegraphics[width=\textwidth]{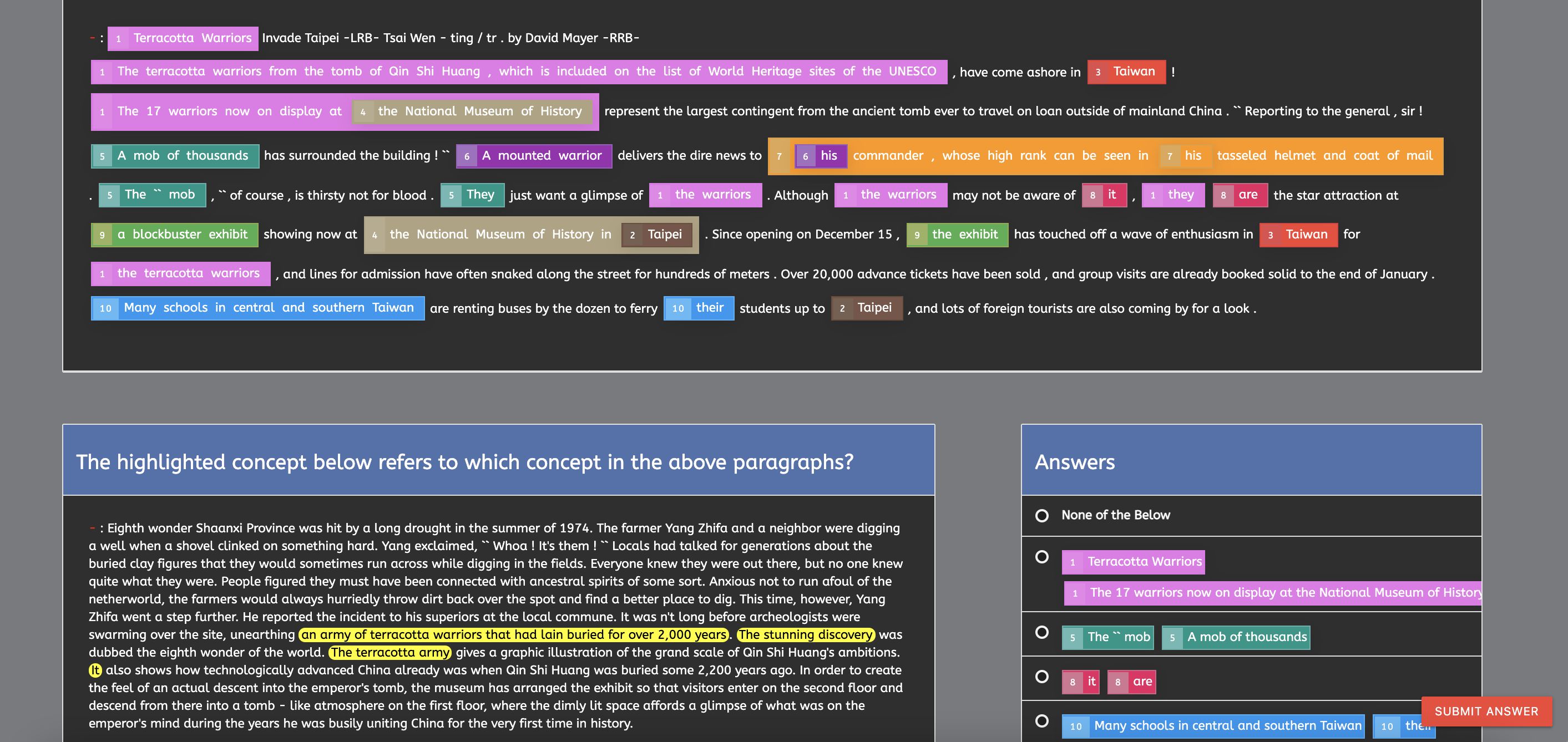}
\label{tool}
\end{figure*}

Figure \ref{tool} shows the annotation tool we created to build \longtonotes .

\begin{figure*}[t]
\caption{The summary page of our annotation tool is shown after all the chain decisions in a paragraph are made. The annotators can look and verify all the decisions and confirm answers and proceed to the next para or can change their answers if they want.}
\includegraphics[width=\textwidth]{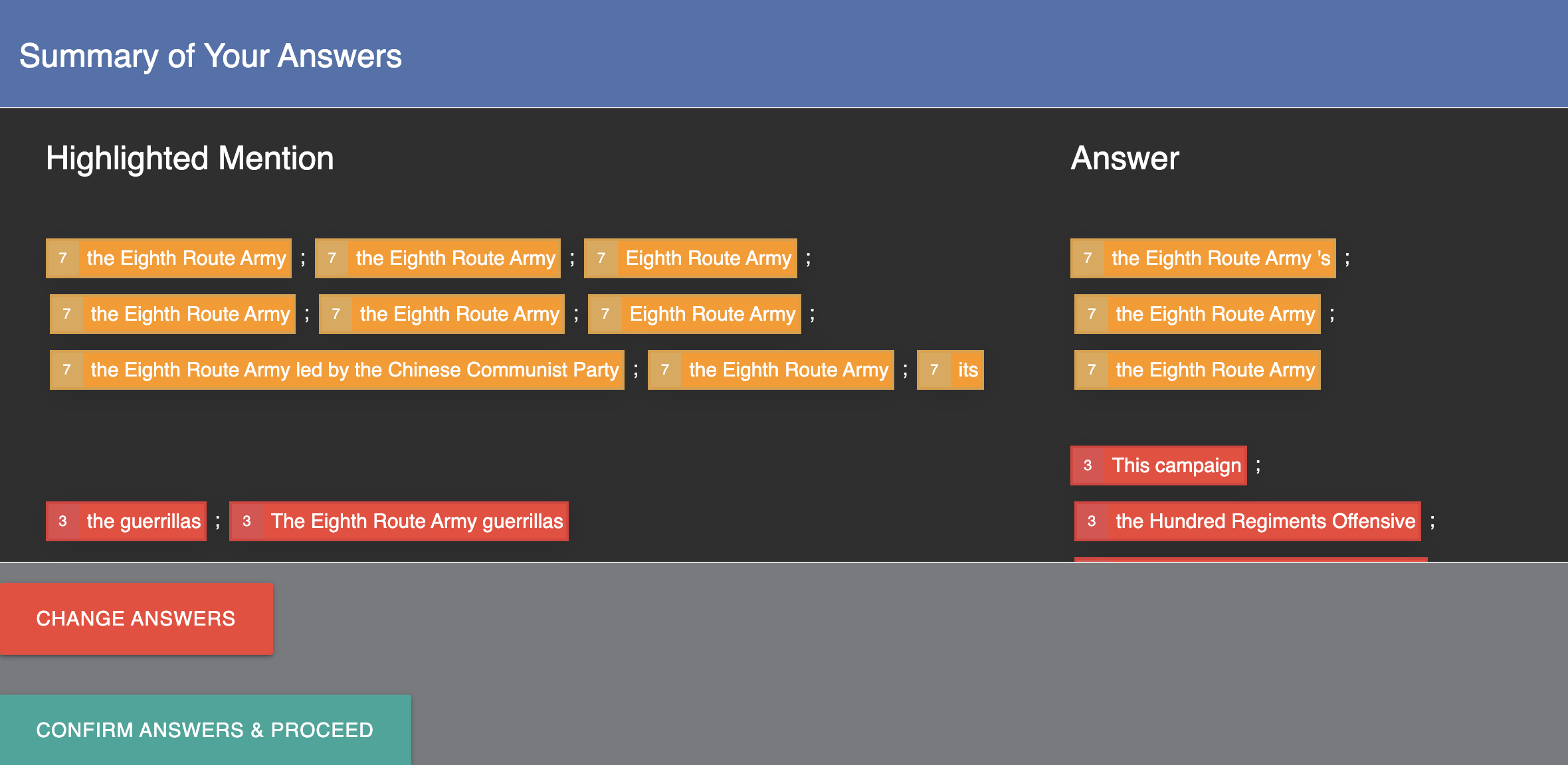}
\label{tool-summary}
\end{figure*}

\subsection{Comparison with OntoNotes}
A detailed genre-wise comparison of the documents from the OntoNotes dataset which were merged in \longtonotes\ is presented in Table \ref{DocComparison}. It can be seen that categories like \texttt{bn} and \texttt{nw} are completely missing in \longtonotes\ , while \texttt{pt} is partially missing.

\begin{table}[h]
\begin{tabular}{ |p{2.0cm}||p{1.5cm}|p{1.5cm}|  }
 \hline
 \multicolumn{3}{|c|}{Documents in Corpus comparison} \\
 \hline
 Category  &  Onto & Longto\\
 \hline
 bc/cctv    & \checkmark &  \checkmark\\
 bc/cnn   & \checkmark   & \checkmark\\
 bc/msnbc  & \checkmark &  \checkmark\\
 bc/phoenix  & \checkmark &  \checkmark\\
 bn/abc   & \checkmark &\xmark\\
 bn/cnn   & \checkmark &\xmark\\
 bn/mnb   & \checkmark &\xmark\\
 bn/nbc   & \checkmark &\xmark\\
 bn/pri   & \checkmark &\xmark\\
 bn/voa   & \checkmark &\xmark\\
 mz/sinorama   & \checkmark & \checkmark\\
 nw/wsj   & \checkmark & \xmark\\
 nw/xinhua   & \checkmark & \xmark\\
 pt/nt & \checkmark & \checkmark \\
 pt/ot & \checkmark & \xmark\\
 tc/ch & \checkmark & \checkmark\\
 wb/a2e   & \checkmark & \checkmark\\
 wb/c2e   & \checkmark & \checkmark\\
 wb/eng  & \checkmark & \checkmark\\
 \hline
 \end{tabular}
 \caption{\textbf{Comparison of documents from various sub-categories that exists in OntoNotes 5.0 and our proposed dataset \longtonotes}} 
 \label{DocComparison}
\end{table}

\subsection{Dataset selection decision}
\label{appendix:selection_discuss}

Due to budget constraints and the expertise of our team and annotators in English only (and some training of annotators is required to ensure data quality), we only considered the English parts of the OntoNotes dataset in our work. We think that the dataset can be extended to Arabic and Chinese too, but we leave it for future work.

\subsection{Annotating singletons}
\label{appendix:singletons_discuss}
While manually annotating all singletons, we observed that almost all NPs can be thought of as mentions and all those NPs that are not part of any chain can be thought of as a singleton. Our analysis suggests that there are over $50\%$ mentions that are not annotated by OntoNotes and can qualify for singletons. To annotate all the singletons, the annotator needs to go through all of them, discard the ones that do not abide by the OntoNotes rules, and then make a decision whether to merge each singleton to some chain or other singleton. In our analysis, the number of such singletons is very low and all the efforts were not worth it for the small improvement over the current annotations. So we decide to ignore all the singletons in our study.

\subsection{Greedy rule-based matching system }
\label{appendix:greedy_discuss}
We use a greedy string matching system where we take all the mentions in a chain of the current para $i+1$ and analyze its part of speech provided in the OntoNotes dataset. We take the first Noun (NN or NP) present in each chain and look for the mentions overlap in all other previous paras $1,\dots,i$ chains. We merged two chains if there is a strict overlap with any of the mentions in a given chain. If there are no strict overlaps, we move to the next noun in the given chain and repeat the process. If we find no strict overlap with any mentions in any other para chains, we keep the chain independent (same as assigning \textit{None of the below} in our annotation tool). We repeat the process with all chains in a given document and constantly update the chain after every para. 

\begin{figure}[h]
\centering
\includegraphics[width=0.5\textwidth]{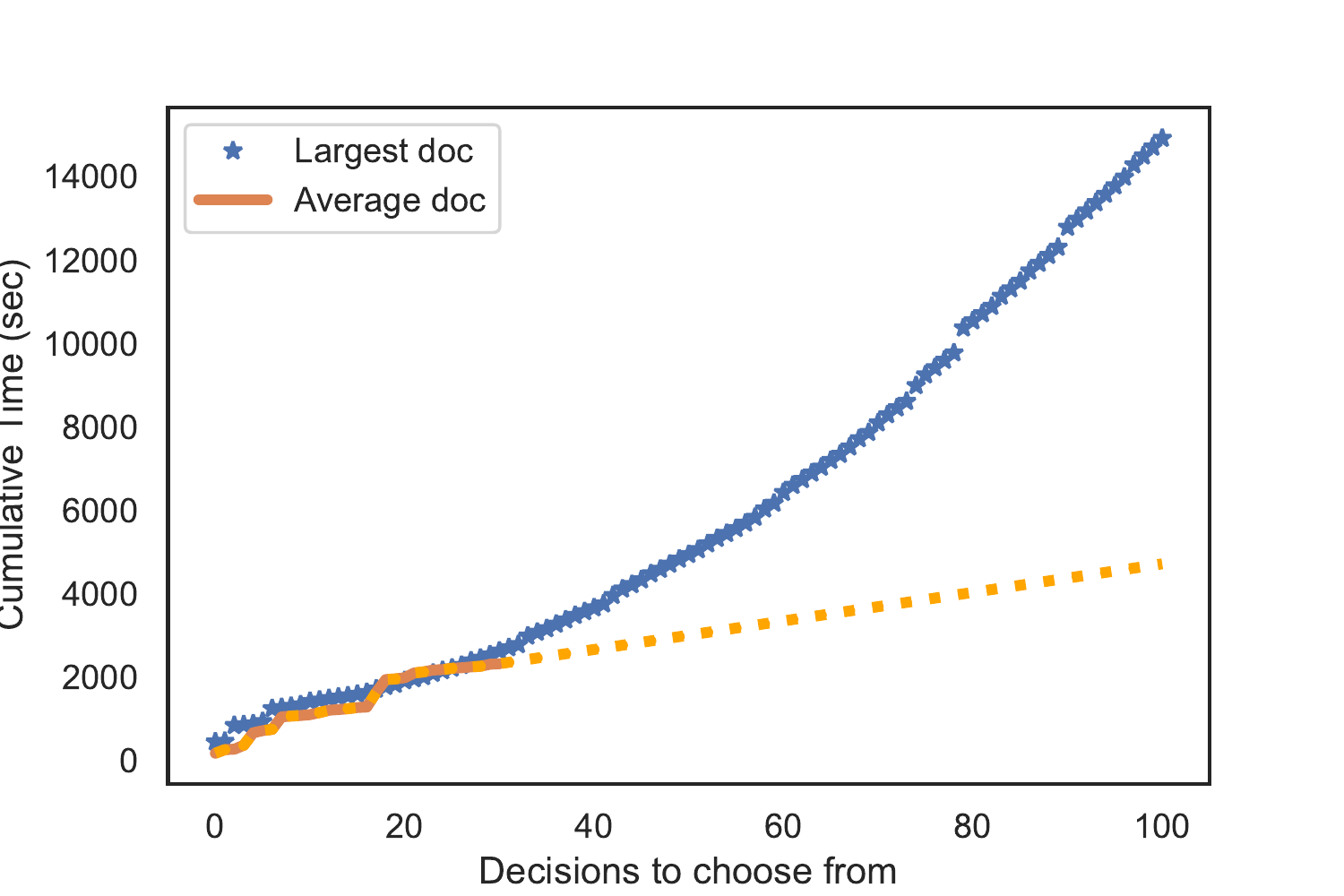}
\caption{\textbf{Annotation Time and Document Length.} Annotation time (cumulative) increases exponentially with the increase in the number of decisions to choose from. A comparison is shown between the longest document in \longtonotes\ vs an average document. The dotted lines represent the increase in annotation time if the growth was linear. }
\label{appendix:TimeAnnotatio}
\end{figure}

\section{Train test dev split}
A comparison between the number of documents in the train-test-dev split between \longtonotes\ and OntoNotes is provided in Table \ref{Train-test-devsplit}.

\begin{table}[h]
\centering
\begin{tabular}{l c c c}
 \toprule
 \bf Dataset  &  Train & Dev & Test \\
 \hline
 OntoNotes    & 2802 &   343 & 348  \\
 \longtonotes   & 1959  & 234 & 222 \\
 
 \bottomrule
 \end{tabular}
 \caption{Comparison of the train-test-dev split of documents between OntoNotes and \longtonotes}

 \label{Train-test-devsplit}
\end{table}

\subsection{Genre-wise disagreement analysis}

Table \ref{genredisagree} presents the genre-wise disagreement analysis for strict decision matching. Genres with longer documents like \texttt{bc, mz} have more disagreements compared to genres with smaller document lengths like \texttt{tc, pt}. 

The trend is very similar for new chain assignments where genres with larger documents have more disagreements over new chain assignments. The numbers are presented in Table \ref{genredisagreenew}.

\begin{figure}[h]
\centering
\includegraphics[width=0.5\textwidth]{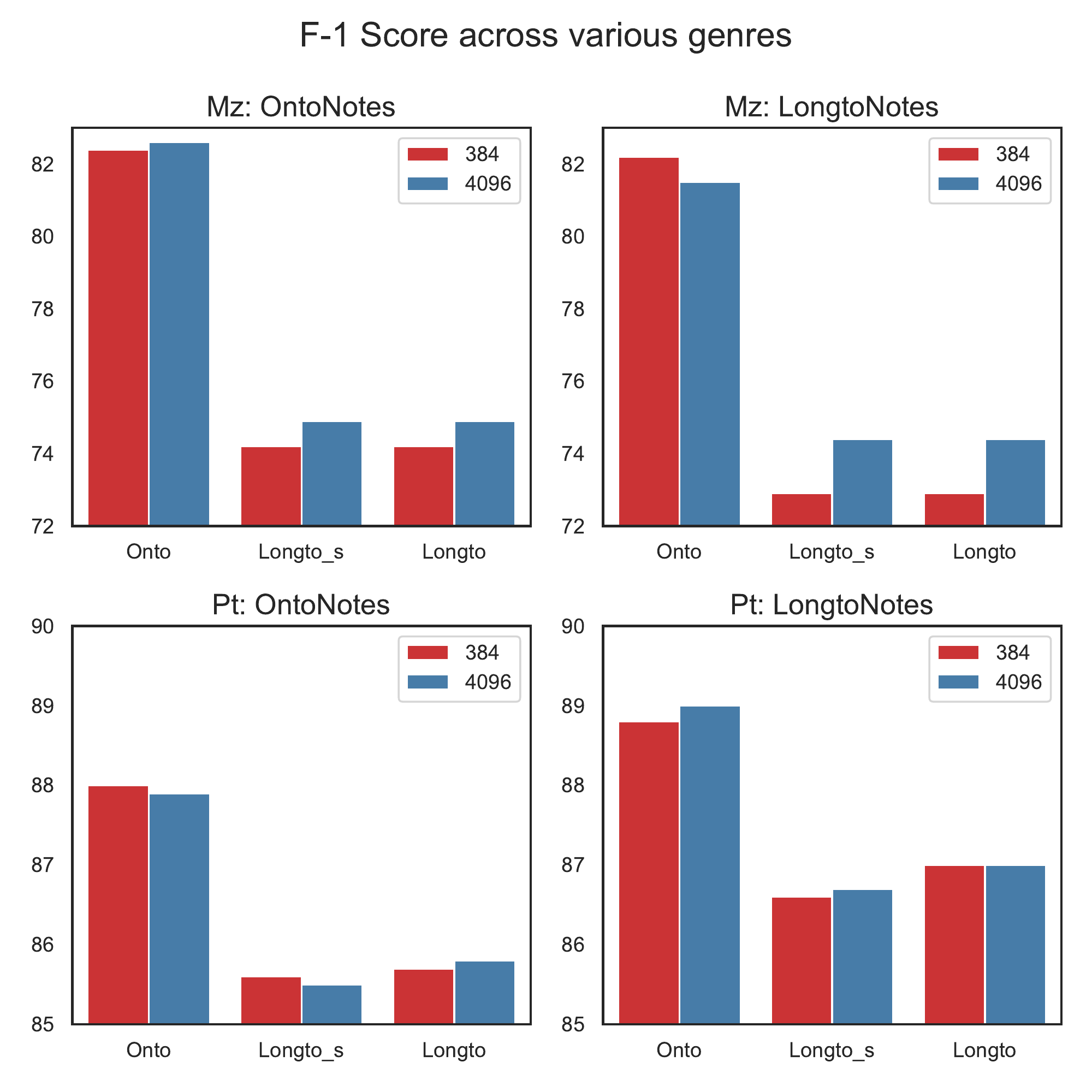}
\caption{Plot comparing the sequence length effect on performance for two genres: \emph{magazine (mz)} and \emph{pivot (pt)}.}
\label{genreF1}
\end{figure}

\begin{table}[h]
\centering
\begin{tabular}{l c c c c}
 \bf   & \multicolumn{2}{c}{\bf bc}& \\
 \hline
 \bf & \textbf{Ann1}  &  \textbf{Ann2} & \textbf{Ann3} \\
 \hline
 \textbf{Ann1}    & 1.0 & 0.91 & 0.87 \\
 \textbf{Ann2}    & 0.91 & 1.0 & 0.88  \\
 \textbf{Ann3}    & 0.87 & 0.88 & 1.0   \\
  \toprule
  \\
 \bf   & \multicolumn{2}{c}{\bf mz}& \\
 \hline
 \bf & \textbf{Ann1}  &  \textbf{Ann2} & \textbf{Ann3} \\
 \hline
 \textbf{Ann1}    & 1.0 & 0.91 & 0.94 \\
 \textbf{Ann2}    & 0.91 & 1.0 & 0.93  \\
 \textbf{Ann3}    & 0.94 & 0.93 & 1.0   \\
 \toprule
 \\
  \bf   & \multicolumn{2}{c}{\bf pt}& \\
 \hline
 \bf & \textbf{Ann1}  &  \textbf{Ann2} & \textbf{Ann3} \\
 \hline
 \textbf{Ann1}    & 1.0 & 0.97 & 0.98 \\
 \textbf{Ann2}    & 0.97 & 1.0 & 0.96  \\
 \textbf{Ann3}    & 0.98 & 0.96 & 1.0   \\
  \toprule
  \\
   \bf   & \multicolumn{2}{c}{\bf tc}& \\
 \hline
 \bf & \textbf{Ann1}  &  \textbf{Ann2} & \textbf{Ann3} \\
 \hline
 \textbf{Ann1}    & 1.0 & 0.99 & 0.98 \\
 \textbf{Ann2}    & 0.99 & 1.0 & 0.98  \\
 \textbf{Ann3}    & 0.98 & 0.98 & 1.0   \\
 \toprule
  \\
   \bf   & \multicolumn{2}{c}{\bf wb}& \\
 \hline
 \bf & \textbf{Ann1}  &  \textbf{Ann2} & \textbf{Ann3} \\
 \hline
 \textbf{Ann1}    & 1.0 & 0.93 & 0.90 \\
 \textbf{Ann2}    & 0.93 & 1.0 & 0.92  \\
 \textbf{Ann3}    & 0.90 & 0.92 & 1.0   \\
 \bottomrule
 \end{tabular}
 \caption{Genre-wise strict decision-based disagreement analysis between the annotators.}
 \label{genredisagree}
\end{table}


\subsubsection{Genre-wise disagreement analysis}
In general, annotators disagree more on pronouns than proper nouns and the trend is consistent for various genres as shown in Table \ref{genrepos}.

\begin{table}[h!]
\centering
\begin{tabular}{l c c }
 \hline
 \bf \textbf{PoS type} & \textbf{bc}  &  \textbf{pt} \\
 \hline
 \textbf{Pronouns}    & 3.6 & 0.04  \\
 \textbf{Nouns}    & 3.2 & 0.05   \\
 \textbf{Proper Nouns}   & 1.9 & 0.03   \\
 \textbf{Verbs}    & 3.5 & 1.0   \\
 \bottomrule
 \end{tabular}
 \caption{Genre wise part of speech comparison for two genres: \texttt{bc} and \texttt{pt}. The numbers are normalized and presented in percentages.}
 \label{genrepos}
\end{table}

\begin{table}[h]
\centering
\begin{tabular}{l c c c c}
 \bf   & \multicolumn{2}{c}{\bf bc}& \\
 \hline
 \bf & \textbf{Ann1}  &  \textbf{Ann2} & \textbf{Ann3} \\
 \hline
 \textbf{Ann1}    & 1.0 & 0.91 & 0.85 \\
 \textbf{Ann2}    & 0.91 & 1.0 & 0.86  \\
 \textbf{Ann3}    & 0.85 & 0.86 & 1.0   \\
  \toprule
  \\
 \bf   & \multicolumn{2}{c}{\bf mz}& \\
 \hline
 \bf & \textbf{Ann1}  &  \textbf{Ann2} & \textbf{Ann3} \\
 \hline
 \textbf{Ann1}    & 1.0 & 0.89 & 0.91 \\
 \textbf{Ann2}    & 0.89 & 1.0 & 0.90  \\
 \textbf{Ann3}    & 0.91 & 0.90 & 1.0   \\
 \toprule
 \\
  \bf   & \multicolumn{2}{c}{\bf pt}& \\
 \hline
 \bf & \textbf{Ann1}  &  \textbf{Ann2} & \textbf{Ann3} \\
 \hline
 \textbf{Ann1}    & 1.0 & 0.94 & 0.95 \\
 \textbf{Ann2}    & 0.94 & 1.0 & 0.91  \\
 \textbf{Ann3}    & 0.95 & 0.91 & 1.0   \\
  \toprule
  \\
   \bf   & \multicolumn{2}{c}{\bf tc}& \\
 \hline
 \bf & \textbf{Ann1}  &  \textbf{Ann2} & \textbf{Ann3} \\
 \hline
 \textbf{Ann1}    & 1.0 & 0.98 & 0.98 \\
 \textbf{Ann2}    & 0.98 & 1.0 & 0.98  \\
 \textbf{Ann3}    & 0.98 & 0.98 & 1.0   \\
 \toprule
  \\
   \bf   & \multicolumn{2}{c}{\bf wb}& \\
 \hline
 \bf & \textbf{Ann1}  &  \textbf{Ann2} & \textbf{Ann3} \\
 \hline
 \textbf{Ann1}    & 1.0 & 0.92 & 0.90 \\
 \textbf{Ann2}    & 0.92 & 1.0 & 0.91  \\
 \textbf{Ann3}    & 0.90 & 0.91 & 1.0   \\
 \bottomrule
 \end{tabular}
 \caption{Genre-wise disagreement analysis between the annotators for new chain assignment.}
 \label{genredisagreenew}
\end{table}


\section{Results}
\label{appendix:empirical}

\subsection{MUC, $B^3$ and CEAFE scores}
Tables \ref{ComparMUC}, \ref{ComparBCUB} and \ref{ComparCEAFE} present the MUC \cite{MUC}, $B^3$ \cite{Bagga98algorithmsfor} and CEAFE \cite{CEAFE} scores for SpanBERT Base \cite{lee-etal-2017-end} and LongDocCoref Models \cite{toshniwal-etal-2020-learning}. On all three metrics, both models trained on \longtonotes\ dataset outperform the models trained on the OntoNotes dataset. For the SpanBERT base model, we compare three versions of the \longtonotes\ dataset: \longtonotessmall\ and \longtonotes\ dataset as mentioned in the paper and \longtonoteseq\ where \longtonotes\ dataset is reweighted to create the total number of documents equal to the number of documents in OntoNotes dataset. For the LongDocCoref model, $n$ represents the maximum number of training sentences, while $m$ refers to the memory used.

\subsection{Genre wise $F_1$ scores vs sequence length}
Table \ref{ComparF1genre384} shows that LongFormer Large model with a larger sequence length (4096) outperforms the one with a shorter sequence length (384) for all models. The difference is higher when the documents are longer (as seen in \texttt{mz} genre) than when the documents are shorter (as seen in \texttt{pt}).

\begin{table}[h!]
\centering
\small
\begin{tabular}{l c c}
 \toprule
 & \multicolumn{2}{c}{\bf Memory Size} \\
 \bf Dataset  &  20 &  40 \\
 \midrule
 OntoNotes & 76.6 & \textbf{77.0}  \\
  \longtonotes & 72.9 &  \textbf{73.7} \\
  \longtonotessmall  & 70.2  & \textbf{70.7} \\
 \bottomrule
 \end{tabular}
 \caption{\textbf{Memory Size vs. Performance}. We compare two settings of the memory size parameter in memory model \cite{toshniwal-etal-2020-learning} and find that the larger memory version achieves better results on each dataset.}
 \label{memorysizeVF1}
\end{table}

\begin{table*}[t]
\resizebox{\textwidth}{!}{
\begin{tabular}{l | ccc | ccc | ccc | ccc | ccc | ccc}
 \toprule
 \bf   & \multicolumn{6}{c}{\bf OntoNotes} & \multicolumn{6}{c}{\bf \longtonotessmall} & \multicolumn{6}{c}{\bf \longtonotes} \\
 & \multicolumn{3}{c}{\bf Mention} & \multicolumn{3}{c}{\bf Coref} & \multicolumn{3}{c}{\bf Mention} & \multicolumn{3}{c}{\bf Coref} & \multicolumn{3}{c}{\bf Mention} & \multicolumn{3}{c}{\bf Coref} \\
  {} &  P & R & $F_1$ & P & R & $F_1$ & P & R & $F_1$ & P & R & $F_1$ & P & R & $F_1$ & P & R & $F_1$ \\
 
  \hline
  LongFormer Large (\texttt{mz})    &  &  &  & &  & &  & &  & &  & & &  & & &  &\\
 \quad + OntoNotes (384)   & 88.0 & 87.9 & 88.0 & 82.4 & 82.4 & 82.4 & 84.3 & 86.1 & 85.2 & 73.8 & 75.0 & 74.2 & 84.3 & 86.1 & 85.2 & 73.8 & 75.0 & 74.2\\
  \quad + OntoNotes (4096)   & 87.9 & 88.3 & 88.1 & 82.4  & 82.9 & \textbf{82.6} & 84.4 & 86.7 & 85.5 & 74.1 & 75.9 & \textbf{74.9} & 84.4 & 86.7 & 85.5 & 74.1 & 75.9 & \textbf{74.9}\\
 \quad +  \longtonotes\ (384)   & 87.0 & 88.4 & 87.7 & 81.4  & 83.0 & \textbf{82.2}  & 84.4 & 86.9 & 85.6 & 72.4 & 73.6 & 72.9 & 84.4 & 86.9 & 85.6 & 72.4 & 73.6 & 72.9\\
  \quad + \longtonotes\ (4096)  & 86.9 & 87.8 & 87.4 & 80.9  & 82.0 & 81.5  & 85.0 & 86.7 & 85.8 & 74.1 & 74.8 & \textbf{74.4} & 85.0 & 86.7 & 85.8 & 74.1 & 74.8 & \textbf{74.4}\\
\\
  \hline
  LongFormer Large (\texttt{pt})    &  &  &  & &  & &  & &  & &  & & &  & & &  &\\
 \quad + OntoNotes (384)   & 95.5 & 94.4 & 95.0 & 88.6 & 87.4 & \textbf{88.0} & 94.3 & 95.3 & 94.8 & 84.6 & 86.9 & 85.7 & 94.9 & 94.4 & 94.7 & 85.5 & 85.8 & \textbf{85.6}\\
  \quad + OntoNotes (4096)   & 95.6 & 94.2 & 94.9 & 88.9  & 86.9 & 87.9 & 94.4 & 94.8 & 94.6 & 84.8 & 86.8 & \textbf{85.8} & 94.9 & 94.0 & 94.5 & 85.5 & 85.2 & 85.5\\
 \quad +  \longtonotes\ (384)   & 95.1 & 94.3 & 94.7 & 89.2 & 88.3 & 88.8  & 94.2 & 95.1 & 94.6 & 86.0 & 88.0 & \textbf{87.0} & 94.6 & 94.2 & 94.4 & 86.5 & 86.7 & 86.6\\
  \quad + \longtonotes\ (4096)  & 95.3 & 94.2 & 94.8 & 89.7 & 88.2 & \textbf{89.0}  & 94.5 & 94.5 & 94.5 & 86.4 & 87.4 & 86.9 & 94.8 & 93.7 & 94.3 & 87.0 & 86.4 & \textbf{86.7}\\
\\
 \bottomrule
 \end{tabular}
 }
 \caption{\textbf{Comparison of \textbf{$F_1$} scores for \texttt{mz} and \texttt{pt} genres.}} 
 \label{ComparF1genre384}
\end{table*}

\begin{table*}[]
\resizebox{\textwidth}{!}{
\begin{tabular}{l | ccc | ccc | ccc | ccc | ccc | ccc}
 \toprule
 \bf   & \multicolumn{6}{c}{\bf OntoNotes} & \multicolumn{6}{c}{\bf \longtonotessmall} & \multicolumn{6}{c}{\bf \longtonotes} \\
 & \multicolumn{3}{c}{\bf Mention} & \multicolumn{3}{c}{\bf Coref} & \multicolumn{3}{c}{\bf Mention} & \multicolumn{3}{c}{\bf Coref} & \multicolumn{3}{c}{\bf Mention} & \multicolumn{3}{c}{\bf Coref} \\
  {} &  P & R & $F_1$ & P & R & $F_1$ & P & R & $F_1$ & P & R & $F_1$ & P & R & $F_1$ & P & R & $F_1$ \\
 \hline
 SpanBERT Base \cite{lee-etal-2017-end}   &  &  &  & &  & &  & &  & &  & & &  & & &  &\\
 \quad + OntoNotes  & 86.6 & 87.5  & 87.0 & 83.1  & 83.6 & 83.4 & 88.4 & 85.0 & 86.7 & 84.2& 80.8 & 82.4 & 86.7  & 85.4  & 86.1  & 83.0  & 81.3  & \textbf{82.1}\\
 \quad + \longtonotessmall & 73.3 & 91.0 & 81.2 & 70.0 & 85.7 & 77.1 & 78.3 & 90.5 & 84.0 & 73.8 & 85.5 & 79.2 & 73.2 & 90.4 & 80.9 & 69.4 & 85.1 & 76.5 \\
 \quad + \longtonotes & 86.6 & 87.1 & 86.8 & 83.0 & 82.9 & \textbf{86.8} & 88.1 & 84.6 & 86.3 & 83.3 & 80.1 & 81.7 & 86.6 & 85.5 & 86.0 & 82.4 & 81.0 & 81.7 \\
 \quad + \longtonoteseq  & 86.1 & 87.8 & 87.0 & 82.8 & 83.5 & 83.2 & 87.7 & 86.2 & 87.0 & 83.4 & 81.9 & \textbf{82.6} &  86.1 & 86.3 & 86.2 & 82.3 & 81.9 & \textbf{82.1}\\
  \\
 LongDocCoref \cite{toshniwal-etal-2020-learning}   &  &  &  & &  & &  & &  & &  & & &  & & &  &\\
 \quad + OntoNotes    & 95.3 & 85.6 & 86.4 & 81.2 & 85.4 & 83.2 & 95.3 & 85.6 & 86.4 & 77.8 & 86.2 & 81.8 & 95.3 & 85.6 & 86.4 & 78.2 & 85.2 & 81.6\\
 \quad + \longtonotessmall  & 95.3 & 85.6 & 86.4 & 22.3 & 66.9 & 33.5 & 95.3 & 85.6 & 86.4 & 17.5 & 65.7 & 27.6 & 95.3 & 85.6 & 86.4 & 21.7 & 66.9 & 32.8\\
 \quad + \longtonotes   & 95.3 & 85.6 & 86.4 & 81.4 & 85.0 & 83.2 & 95.3 & 85.6 & 86.4 & 79.3 & 85.8 & 82.4 & 95.3 & 85.6 & 86.4 & 79.1 & 85.0 & 81.9\\
 \quad + \longtonoteseq\ (n=3)  & 95.3 & 85.6 & 86.4 & 81.6 & 85.2 & \textbf{83.4} & 95.3 & 85.6 & 86.4  & 79.7 & 86.2 & \textbf{82.8} & 95.3 & 85.6 & 86.4 & 79.3 & 85.2 & \textbf{82.2}\\
 \quad + \longtonoteseq\ (n=5)  & 95.3 & 85.6 & 86.4 & 81.4 & 85.3 & 83.3 &95.3 & 85.6 & 86.4  & 79.7 & 86.2 & \textbf{82.8}  & 95.3 & 85.6 & 86.4 & 79.2 & 85.3 & 82.1\\
  \quad + \longtonoteseq\ (n=10)  & 95.3 & 85.6 & 86.4 & 81.5 & 85.1 & 83.3 & 95.3 & 85.6 & 86.4  & 79.7 & 86.2 & \textbf{82.8}  & 95.3 & 85.6 & 86.4 & 79.6 & 84.8 & 82.1\\
 \quad + \longtonoteseq\ (n=10, m=40)  & 95.3 & 85.6 & 86.4 & 81.6 & 85.6 & 83.6 & 95.3 & 85.6 & 86.4  & 79.8 & 85.9 & 82.7 & 95.3 & 85.6 & 86.4 & 79.5 & 85.2 & 82.3\\
 \\
 \bottomrule
 \end{tabular}
 }
 \caption{\textbf{Comparison of \textbf{MUC} scores}} 
 \label{ComparMUC}

\end{table*}

\begin{table*}[]
\resizebox{\textwidth}{!}{
\begin{tabular}{l | ccc | ccc | ccc | ccc | ccc | ccc}
 \toprule
 \bf   & \multicolumn{6}{c}{\bf OntoNotes} & \multicolumn{6}{c}{\bf \longtonotessmall} & \multicolumn{6}{c}{\bf \longtonotes} \\
 & \multicolumn{3}{c}{\bf Mention} & \multicolumn{3}{c}{\bf Coref} & \multicolumn{3}{c}{\bf Mention} & \multicolumn{3}{c}{\bf Coref} & \multicolumn{3}{c}{\bf Mention} & \multicolumn{3}{c}{\bf Coref} \\
  {} &  P & R & $F_1$ & P & R & $F_1$ & P & R & $F_1$ & P & R & $F_1$ & P & R & $F_1$ & P & R & $F_1$ \\
 \hline
 SpanBERT Base \cite{lee-etal-2017-end}   &  &  &  & &  & &  & &  & &  & & &  & & &  &\\
 \quad + OntoNotes  & 86.6 & 87.5  & 87.0 & 75.0 & 75.5 & \textbf{75.3}  & 88.4 & 85.0 & 86.7 & 70.7& 65.1 & 67.8 &  86.7 & 85.4 & 86.1 & 72.3 & 69.5  & 70.9\\
 \quad + \longtonotessmall & 73.3 & 91.0 & 81.2  & 57.0 & 76.8  & 65.4 & 78.3 & 90.5 & 84 & 54.8 & 69.7 & 61.3 & 73.2 & 90.4 & 80.9 & 53.3 & 72.8 & 61.5 \\
 \quad + \longtonotes & 86.6 & 87.1 & 86.8 & 74.6 & 74.0 & 74.3 & 88.1 & 84.6 & 86.3 & 67.5 & 62.7 & 65.0 & 86.6 & 85.5 & 86.0 & 70.6 & 68.2 & 69.4 \\
 \quad + \longtonoteseq  & 86.1 & 87.8 & 87.0 & 74.9 & 75.2 & 75.0 & 87.7 & 86.2 & 87.0 & 69.7 & 67.0 & \textbf{68.3} & 86.1 & 86.3 & 86.2 & 71.7 & 70.6 & \textbf{71.2}\\
  \\
 LongDocCoref \cite{toshniwal-etal-2020-learning}    &  &  &  & &  & &  & &  & &  & & &  & & &  &\\
 \quad + OntoNotes    & 95.3 & 85.6 & 86.4 & 72.2 & 77.9 & 74.9 & 95.3 & 85.6 & 86.4 & 57.9 & 71.7 & 64.0 & 95.3 & 85.6 & 86.4 & 63.9 & 74.7 & 68.9\\
 \quad + \longtonotessmall   & 95.3 & 85.6 & 86.4 & 18.3 & 61.7 & 28.2 & 95.3 & 85.6 & 86.4 & 10.7 & 53.6 & 17.9  & 95.3 & 85.6 & 86.4 & 16.1 & 58.7 & 25.2\\
 \quad + \longtonotes   & 95.3 & 85.6 & 86.4 & 73.3 & 76.7 & 75.0  & 95.3 & 85.6 & 86.4 & 61.0 & 70.1 & 65.2 & 95.3 & 85.6 & 86.4 & 65.5 & 73.7 & 69.4\\
 \quad +\longtonoteseq\ (n=3)  & 95.3 & 85.6 & 86.4 & 73.7 & 76.9 & 75.2 &95.3 & 85.6 & 86.4  & 64.4 & 70.4 & 67.3 & 95.3 & 85.6 & 86.4 & 67.5 & 73.7 & 70.5\\
 \quad + \longtonoteseq\ (n=5)  & 95.3 & 85.6 & 86.4 & 73.4 & 77.3 & 75.3 &95.3 & 85.6 & 86.4  & 64.5 & 70.9 & \textbf{67.6} & 95.3 & 85.6 & 86.4 & 67.5 & 74.2 & 70.7\\
 \quad + \longtonoteseq\ (n=10)  & 95.3 & 85.6 & 86.4 & 73.6 & 77.0 & 75.3 &95.3 & 85.6 & 86.4  & 64.5 & 70.9 & \textbf{67.6} & 95.3 & 85.6 & 86.4 & 68.3 & 73.5 & 70.8\\
 \quad + \longtonoteseq\ (n=10, m=40)  & 95.3 & 85.6 & 86.4 & 73.5 & 78.1 & \textbf{75.7} & 95.3 & 85.6 & 86.4  & 65.0 & 70.5 & \textbf{67.6} & 95.3 & 85.6 & 86.4 & 67.9 & 74.4 & \textbf{71.0}\\
 \\
 \bottomrule
 \end{tabular}
 }
 \caption{\textbf{Comparison of \textbf{BCUB} scores}} 
 \label{ComparBCUB}

\end{table*}

\begin{table*}[h]
\resizebox{\textwidth}{!}{
\begin{tabular}{l | ccc | ccc | ccc | ccc | ccc | ccc}
 \toprule
 \bf   & \multicolumn{6}{c}{\bf OntoNotes} & \multicolumn{6}{c}{\bf \longtonotessmall} & \multicolumn{6}{c}{\bf \longtonotes} \\
 & \multicolumn{3}{c}{\bf Mention} & \multicolumn{3}{c}{\bf Coref} & \multicolumn{3}{c}{\bf Mention} & \multicolumn{3}{c}{\bf Coref} & \multicolumn{3}{c}{\bf Mention} & \multicolumn{3}{c}{\bf Coref} \\
  {} &  P & R & $F_1$ & P & R & $F_1$ & P & R & $F_1$ & P & R & $F_1$ & P & R & $F_1$ & P & R & $F_1$ \\
 \hline
 SpanBERT Base \cite{lee-etal-2017-end}  &  &  &  & &  & &  & &  & &  & & &  & & &  &\\
 \quad + OntoNotes  & 86.6 & 87.5  & 87.0 & 71.5 & 73.7 & \textbf{72.1}  & 88.4 & 85.0 & 86.7 & 63.3& 61.6 & 62.4 & 86.7 & 85.4 & 86.1 & 68.1 & 68.4  & 68.2\\
 \quad + \longtonotessmall & 73.3 & 91.0  & 81.2  & 53.2 & 69.5  & 60.3 & 78.3 & 90.5 & 84.0 & 51.5 & 59.2 & 55.1 & 73.2 & 90.4 & 80.9 & 50.4 & 64.2 & 56.5 \\
 \quad + \longtonotes & 86.6 & 87.1 & 86.8 & 70.8 & 73.1 & 71.9 & 88.1 & 84.6 & 86.3 & 63.4 & 60.5 & 61.9 & 86.6 & 85.5 & 86.0 & 67.7 & 68.2 & 67.9 \\
 \quad +\longtonoteseq  & 86.1 & 87.8 & 87.0 & 70.2 & 74.2 & \textbf{72.1} & 87.7 & 86.2 & 87.0 & 64.0 & 63.1 & \textbf{63.5} & 86.1 & 86.3 & 86.2 & 67.5 & 69.6 & \textbf{68.5}\\
  \\
 LongDocCoref \cite{toshniwal-etal-2020-learning}    &  &  &  & &  & &  & &  & &  & & &  & & &  &\\
 \quad + OntoNotes    & 95.3 & 85.6 & 86.4 & 67.0 & 74.5 & 70.5 & 95.3 & 85.6 & 86.4 & 54.5 & 63.4 & 58.6& 95.3 & 85.6 & 86.4 & 61.6 & 69.8 & 65.4\\
 \quad + \longtonotessmall   & 95.3 & 85.6 & 86.4 & 25.7 & 60.0 & 35.9 & 95.3 & 85.6 & 86.4 & 16.8 & 47.8 & 24.8  & 95.3 & 85.6 & 86.4 & 23.5 & 57.2 & 33.3\\
 \quad + \longtonotes   & 95.3 & 85.6 & 86.4 & 65.8 & 75.3 & 70.2 & 95.3 & 85.6 & 86.4 & 53.7 & 65.9 & 59.2 & 95.3 & 85.6 & 86.4 &  60.5 & 71.7 & 65.6\\
 \quad + \longtonoteseq\ (n=3)  & 95.3 & 85.6 & 86.4 & 66.1 & 76.2 & 70.8 &95.3 & 85.6 & 86.4  & 54.9 & 67.4 & 60.5 & 95.3 & 85.6 & 86.4 & 61.2 & 72.2 & 66.2\\
 \quad + \longtonoteseq\ (n=5)  & 95.3 & 85.6 & 86.4 & 66.7 & 76.0 & 71.1  &95.3 & 85.6 & 86.4  & 56.0 & 66.6 & 60.9 & 95.3 & 85.6 & 86.4 & 61.9 & 71.8 & 66.5\\
 \quad +\longtonoteseq\ (n=10)  & 95.3 & 85.6 & 86.4 & 66.2 & 75.9 & 70.7  &95.3 & 85.6 & 86.4  & 56.0 & 66.6 & 60.9 & 95.3 & 85.6 & 86.4 & 61.7 & 72.2 & 66.6\\
 \quad + \longtonoteseq\ (n=10, m=40)  & 95.3 & 85.6 & 86.4 & 68.0 & 75.9 & \textbf{71.7}  &95.3 & 85.6 & 86.4  & 56.1 & 68.9 & \textbf{61.9} & 95.3 & 85.6 & 86.4 & 62.9 & 72.9 & \textbf{67.5}\\
 \\
 \bottomrule
 \end{tabular}
 }
 \caption{\textbf{Comparison of \textbf{CEAFE} scores}} 
 \label{ComparCEAFE}
\end{table*}